\documentclass[twoside]{article}

\usepackage{arxiv}
\usepackage[round]{natbib}

\usepackage{graphicx}
\usepackage{dsfont}
\usepackage{tikz}
\usetikzlibrary{shapes.geometric}
\usepackage{hyperref}
\usepackage{cancel}
\usepackage{caption}

\usepackage{mathrsfs}
\usepackage{amsmath,amssymb,amsthm,enumitem,amsfonts, nccmath}
\usepackage{bbm}
\usepackage{mathtools}
\usepackage{booktabs}
\newtheorem*{theorem*}{Theorem}

\theoremstyle{plain}

\theoremstyle{definition}
\usepackage{algorithm}
\usepackage{algpseudocode}

\theoremstyle{remark}

\newtheoremstyle{new}
  {\topsep} 
  {\topsep} 
  {} 
  {} 
  {\bfseries} 
  {.} 
  {.5em} 
  {} 

\usepackage{enumitem}
\setlist{nosep}

\theoremstyle{new}
\newtheorem{Theorem}{Theorem}

\newtheorem{Definition}{Definition}

\usepackage{nicefrac}
\usepackage{etoolbox}
\usepackage{xcolor}
\newcommand{\bluecolor}[1]{\textcolor{blue}{#1}}
\newcommand{\redcolor}[1]{\textcolor{red}{#1}}
\newcommand{\greencolor}[1]{\textcolor{green}{#1}}

\usepackage{amsmath,amsfonts,bm}
\usepackage{mathrsfs}
\usepackage{algorithm,algpseudocode}

\def\eqref#1{equation~\ref{#1}}

\def\1{\bm{1}}

\DeclareMathAlphabet{\mathsfit}{\encodingdefault}{\sfdefault}{m}{sl}
\SetMathAlphabet{\mathsfit}{bold}{\encodingdefault}{\sfdefault}{bx}{n}

\DeclareMathOperator*{\argmin}{arg\,min}

\newenvironment{tightemize}
{\begin{itemize}\itemsep1pt \parskip0pt \parsep0pt}{\end{itemize}\vspace{-\topsep}}
\DeclareMathSymbol{\shorthyphen}{\mathord}{AMSa}{"39}

\title{Semiparametric conformal prediction}

\author{Ji~Won Park \\
Prescient Design, Genentech \\
\And Robert Tibshirani \\
Department of Statistics,
Stanford University \\
\AND Kyunghyun Cho \\
Prescient Design, Genentech \\
Center for Data Science, New York University
}
\renewcommand{\headeright}{Park, Tibshirani, and Cho}
\renewcommand{\shorttitle}{Semiparametric conformal prediction}
\renewcommand{\undertitle}{}

\begin{document}

%

%

\maketitle
\begin{abstract}
    Many risk-sensitive applications require well-calibrated prediction sets over multiple, potentially correlated target variables, for which the prediction algorithm may report correlated errors.
    In this work, we aim to construct the conformal prediction set accounting for the joint correlation structure of the vector-valued non-conformity scores.
    Drawing from the rich literature on multivariate quantiles and semiparametric statistics, we propose an algorithm to estimate the $1-\alpha$ quantile of the scores, where $\alpha$ is the user-specified miscoverage rate.
    In particular, we flexibly estimate the joint cumulative distribution function (CDF) of the scores using nonparametric vine copulas and improve the asymptotic efficiency of the quantile estimate using its influence function. The vine decomposition allows our method to scale well to a large number of targets.
    As well as guaranteeing asymptotically exact coverage, our method yields desired coverage and competitive efficiency on a range of real-world regression problems, including those with missing-at-random labels in the calibration set.
\end{abstract}

\section{INTRODUCTION} \label{sec:intro}
Many problems in industrial and pharmaceutical applications can be framed as multi-target regression, where the goal is to predict a vector target taking values in $\mathbb{R}^d$ from some input features. In early-stage drug discovery, for example, it is necessary to jointly characterize the ADME (absorption, distribution, metabolism, and excretion) properties of a candidate molecule. The complex biophysical processes governing these properties manifest as correlations in their measurements, often pronounced in the tails \citep{jain2017biophysical,niepel2019multi,du2023admet}. 

\begin{figure}[t!]
\centering
\includegraphics[width=0.7\linewidth]{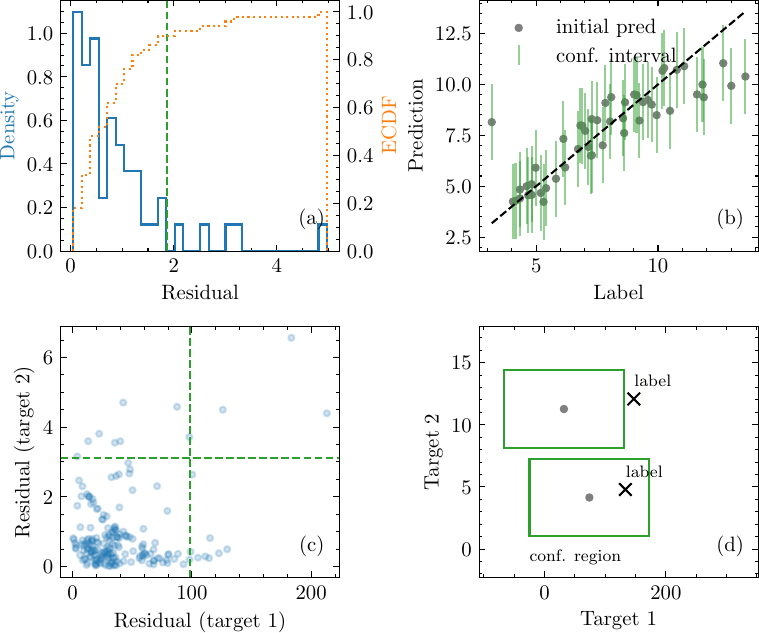}
\caption{\small Top: Single-target split CP. Green dashed line in (a) marks the $1-\alpha$ quantile of calibration scores, setting the confidence interval around test predictions in (b). Bottom: Independent split CP for two targets, with $\sqrt{1-\alpha}$ quantiles marked by green dashed lines in (c). Large prediction sets for two test instances shown in (d). \label{fig:cp_intro}}
\vspace{-0.5cm}
\end{figure}

We often require \emph{prediction sets} capturing uncertainties, rather than point estimates \citep{eklund2015application,cortes-ciriano2020concepts}. The medicinal chemists may use the uncertainties on a given molecule to decide whether to test it in the wet lab. One way to express uncertainties is as confidence at a specified coverage level. Conformal prediction (CP) stands out as a powerful framework that can wrap around any underlying prediction algorithm to produce valid confidence sets for any data distribution \citep{gammerman1998learning,saunders1999transduction,vovk2005algorithmic}. 
A confidence set at the $1-\alpha$ level for a random variable $Y$, denoted ${\Gamma}_{1-\alpha}$, is valid if it contains the true value of $Y$ at least $1-\alpha$ of the time, that is, $\mathbb{P}[Y \in {\Gamma}_{1-\alpha}] \geq 1-\alpha$.

Central to the CP algorithm is the concept of ranks and quantiles \citep{kuchibhotla2020exchangeability}. The confidence set at level $1-\alpha$ is constructed by extracting the $1-\alpha$ quantile of the \emph{non-conformity score} quantifying how much a prediction deviates from the labels. In the context of split CP \citep{papadopoulos2002inductive,lei2015conformal} for single-target regression, for instance, one may define the score as the absolute error and rank the scores evaluated on the calibration set to obtain the $1-\alpha$ quantile, as illustrated in \autoref{fig:cp_intro}(a, b).

CP can be applied independently for each target, not accounting for the correlation in the scores \citep{papadopoulos2002inductive}. Ranking then proceeds like the single-target case at a marginal level $1 - \alpha_j>1 - \alpha$ for each target $j$ that would satisfy the desired global coverage level of $1-\alpha$. By treating the scores as independent, however, we miss out on potential gains in efficiency, or the tightness of the prediction \citep{neeven2018conformal}. \autoref{fig:cp_intro}(c, d) illustrates independent CP for two-target regression. Setting $1-\alpha_j = \sqrt{1-\alpha}$ marginally results in large confidence sets. In general, the calibration set may not be large enough to accommodate extreme $1-\alpha_j$ levels, forced to be closer to 1 as $d$ increases.

\begin{figure}[t!]
\centering
\includegraphics[width=0.7\linewidth]{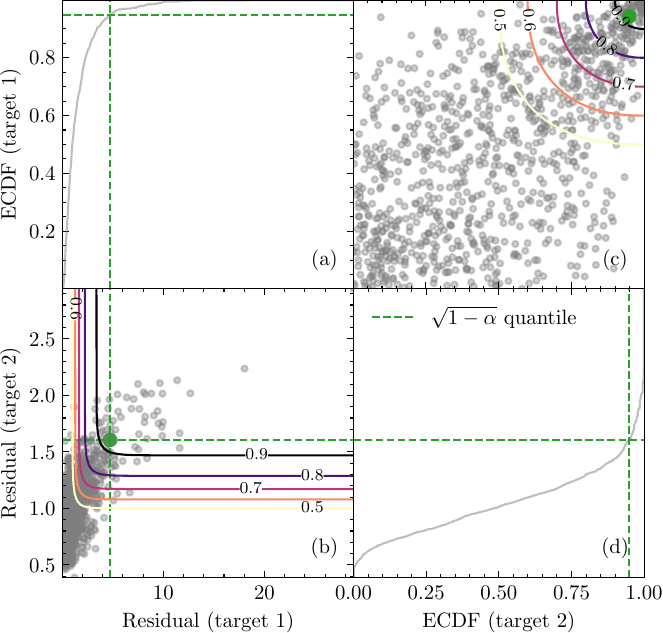}
\caption{\small Scores with an upper tail correlation and long-tailed marginals. The joint 0.9-quantile of the scores can yield more efficient prediction sets than the independent $\sqrt{0.9}$ quantiles of the scores (green dot). (a,~d): Marginal empirical CDF of the scores, from which the $\sqrt{0.9}$ quantiles are extracted. (b): Scores overlaid with level curves of the joint CDF, with the black curve representing the $0.9$ level. (c): Version of (b) viewed in the copula space. \label{fig:joint_quantile}}
\end{figure}

We instead model the targets jointly by viewing the target-wise scores as random vectors $S$ that take values in $\mathbb{R}^d$. The key question then is how to define the $1-\alpha$ quantile of observed $S$. While ranks and quantiles in $\mathbb{R}$ form the basis of nonparametric statistics \citep{lehmann1975statistical}, a canonical ordering does not exist in $\mathbb{R}^d$ for $d>1$. Several notions of multivariate quantiles have been proposed, including ones based on statistical depth \citep{liu1990notion,zuo2003projection} and geometry \citep{chaudhuri1996geometric,hallin2010multivariate}. We build on the work of \citet{messoudi2021copula} that proposed using the \emph{copula}, a statistical tool for modeling the dependence structure of multivariate random variables \cite{nelsen2006introduction}. Copulas can be used to estimate the joint cumulative distribution function (CDF) of the scores, $F(s) = \mathbb{P}[S_1 \leq s_1, \dotsc, S_d \leq s_d]$, where $s \in \mathbb{R}^d$. The quantile is its generalized inverse, or the $p$-level curve $F^{-1}(p) = \{s \in \mathbb{R}^d: F(s) = p\}$. \autoref{fig:joint_quantile} depicts the level curves of the copula (colored contours) fit on correlated scores (gray dots). For 
$1-\alpha=0.9$, the $0.9$-level line contains quantiles with smaller norms compared to the independent target approach (green dot), indicating that joint modeling can yield more efficient prediction sets.

The main novelty of our approach lies in viewing the task of determining $F^{-1}(1-\alpha)$ as a semiparametric inference problem. Our goal is to flexibly estimate $F$ using a nonparametric model while ensuring that inference remains robust and precise for a particular estimand of interest, namely the quantile $F^{-1}(1-\alpha)$. 

As an update to prior work, we estimate $F$ using vine copulas \citep{bedford2002vines}, specifically the nonparametric variant \citep{nagler2017nonparametric} for additional expressibility. Vine copulas can scale well to many targets, as they factorize high-dimensional densities into a product of bivariate densities. Applications of copulas for CP have previously been limited to the empirical copula or a small set of parametric families, which can be biased when observations are sparse or the model misspecified \citep{messoudi2021copula,sun2024copula}. We also demonstrate that a particular copula model allows working with missing-at-random (MAR) target labels in the calibration set.

Having inferred $F$, rather than simply using the plug-in estimate of the quantile, we tailor our analysis to this specific functional by computing the one-step estimator \citep{le1956asymptotic} based on its \emph{influence function} \citep{hampel1974influence}, best understood as a generalized gradient of the functional with respect to small perturbations of $F$. This debiased estimator specifically targets the quantile estimand of interest and carries better theoretical properties like $\sqrt{n}$-consistency and asymptotic normality \citep{hines2022demystifying}. 

{\bf Contributions:} (1) We propose the use of nonparametric vine copulas for flexibly estimating the joint distribution of the non-conformity scores in multivariate conformal calibration (\autoref{sec:density_estimation}). (2) We propose the semiparametric one-step estimator for debiasing the $1{-}\alpha$ score quantile used to construct the prediction sets (\autoref{sec:semiparametric_correction}). (3) We prove that our method produces prediction sets with asymptotically exact coverage and present a variant with finite-sample distribution-free validity. (4) We empirically demonstrate the validity and efficiency of our method on various real-world regression problems (\autoref{sec:real_data}). 

\section{RELATED WORK}
A large body of work aims at extending conformalized quantile regression \citep{romano2019conformalized} to multiple targets \citep{kong2012quantile,paindaveine2011directional,feldman2023calibrated}. There are parallel efforts to make full CP more computationally efficient for some classes of multi-target predictors \citep{johnstone2022exact,nouretdinov2001ridge}. Our focus is to improve the calibration step without additional assumptions on the underlying algorithm. 

When there are multiple response variables, a common approach is to design a single scalar score, such as the $L_2$ distance between the vector-valued prediction and the label  \citep{kuleshov2018conformal,johnstone2021conformal,messoudi2022ellipsoidal,feldman2023calibrated}. This implies choosing a target aggregation scheme, however, and the resulting score is sensitive to the relative scales in the predictions and labels across targets. Moreover, the labels for all $d$ targets need to be observed for score evaluation, which may be impractical for large $d$. 

Another approach is to conformalize the predictions for each individual target \citep{neeven2018conformal}. Non-conformity scores are evaluated target-wise and the score at the $1-\alpha_j$ quantile extracted for each target $j$ such that $\prod_j (1-\alpha_j) = 1-\alpha$. If all targets are to be treated equally, we set $1-\alpha_j = (1-\alpha)^{1/d} > 1-\alpha$ for all $j$. Note that as $d$ increases, $(1-\alpha)^{1/d}$ approaches $1$, leading to inefficient prediction sets. Such extreme levels also require the calibration dataset to be sufficiently large.  

If the scores are highly correlated, we can exploit the dependence structure to improve efficiency. \citet{messoudi2021copula} pioneered the use of copulas for jointly modeling the distribution of vector scores. \citet{sun2024copula} applied this idea to multi-step time series forecasting. The empirical copulas used in previous work can be unstable with sparse observations, especially in the upper tails near the quantile level of interest. \citet{messoudi2021copula} also uses simple parametric copulas like Gumbel and Gaussian, which offer a smoother fit but can suffer from model misspecification. Copulas are reviewed in \autoref{sec:vine_copulas}.

\section{Problem setup} \label{sec:problem_setup}
Consider a multivariate regression task given a dataset $\{(X^{(i)}, Y^{(i)})\}_{i \in \mathcal{I}}$ of input features $X^{(i)} \in \mathcal{X}$ and vector-valued labels $Y^{(i)} \in \mathbb{R}^d$. The dataset can be viewed as $|\mathcal{I}|$ exchangeable samples drawn from the joint distribution $P_{XY} = P_X \times P_{Y|X}$. Given a new test point $X^*$ and a prespecified miscoverage level $\alpha \in [0, 1)$, we would like to determine the set $ {\Gamma}_{1-\alpha}(X^*) \subset \mathbb{R}^d$ that has valid coverage.
\begin{Definition}[Marginal validity]
    A set $\Gamma_{1-\alpha}$ is \emph{marginally valid}\footnote{This is a weaker condition than \emph{conditional validity}, $\mathbb{P}[Y^* \in {\Gamma}_{1-\alpha}(X^*)|X^*] \geq 1-\alpha$.} at level $1-\alpha$ if it contains the response $Y^*$ with probability at least $1-\alpha$:
\begin{align} \label{eq:marginal_validity}
    \mathbb{P}[Y^* \in {\Gamma}_{1-\alpha}(X^*)] \geq 1-\alpha.
\end{align}
\end{Definition} 
Suppose an algorithm ${\hat f}: \mathcal{X} \rightarrow \mathbb{R}^d$ predicts the multiple responses from the input. Our goal is to equip this model, henceforward referred to as the ``underlying'' algorithm, with valid confidence sets. We assume ${\hat f}$ to be a point predictor throughout the main text, and \autoref{app:underlying_predictor} includes results for a conditional density estimator ${\hat f}: \mathcal{X} \rightarrow \mathbb{R}^d \times \mathbb{R}^d_+$ that outputs the mean and variance of the Gaussian predictive distribution.

\section{BACKGROUND} \label{sec:background}

\paragraph{Conformal prediction (CP)} constructs prediction sets without any assumptions on the data distribution $P_{XY}$ \citep{vovk2005algorithmic}. The set is guaranteed to be marginally valid for finite samples as long as the data are exchangeable. CP was originally formulated in the transductive setting \citep{gammerman1998learning,saunders1999transduction,vovk2005algorithmic}. Here we focus on split (inductive) CP \citep{papadopoulos2002inductive,lei2018distribution}, where the labeled dataset $\{(X^{(i)}, Y^{(i)})\}_{i \in \mathcal{I}}$ with index set $\mathcal{I}$ is split into the proper training data $\mathcal{I}_{\rm train}$, used to fit the underlying algorithm ${\hat f}$, and the calibration data $\mathcal{I}_{\rm cal}$, used to conformalize it.\footnote{We present our framework in terms of split CP for clarity, but it applies to full (transductive) \citep{vovk2005algorithmic} and CV \citep{vovk2015cross}, jackknife \citep{barber2021predictive} variants as well.}

Calibration involves the \emph{non-conformity score}, denoted $V(X, Y, {\hat f})$, that measures the extent to which a query point $(X, Y)$ deviates from the patterns learned by $\hat f$ from the training set. Higher values indicate more disagreement. 

Let us first consider the $d=1$ regression case and define the score as the absolute error, $V(X^{(i)}, Y^{(i)}, {\hat f}) = |{\hat f}(X^{(i)}) - Y^{(i)}| \eqcolon S^{(i)} \in \mathbb{R}_+$. Given a miscoverage rate $\alpha \in [0, 1]$ and a test input $X^* \in \mathcal{X}$, the conformal set gathers outcomes yielding the smallest scores:
\begin{align} \label{eq:conf_set}
{\Gamma}_{1-\alpha}(X^*) = \big \{Y \in \mathbb{R}: V(X^*, Y, {\hat f}) \leq Q_{1-\alpha} \big \}, 
\end{align}
where $Q_{1-\alpha}$ is the $\lceil (1-\alpha)(n+1) \rceil$-th smallest of the scores evaluated on the calibration set, $\{S^{(i)}\}$, which is of size $n = |\mathcal{I}_{\rm cal}|$. Equivalently, 
\begin{align} \label{eq:cdf_form}
     {\hat F}(Q_{1-\alpha}) &= \frac{\lceil (1-\alpha)(n+1) \rceil}{n + 1},
\end{align}
where ${\hat F}(s) = \frac{1}{n}\sum_{i \in \mathcal{I}_{\rm cal}} \mathds{1}[S^{(i)} \leq s]$ is the empirical CDF. Applying \autoref{eq:conf_set} yields the confidence interval $[{\hat f}(X^*) - Q_{1-\alpha}, {\hat f}(X^*) + Q_{1-\alpha}]$. We refer readers to \citep{shafer2008tutorial,angelopoulos2021gentle} for more details.

For $d>1$, we may adopt the target-wise absolute error as the score:
\begin{align} \label{eq:score_targetwise}
    V(X^{(i)}, Y^{(i)}, {\hat f})_j = |{\hat f}(X_j^{(i)}) - Y_j^{(i)}| \eqcolon S_j^{(i)},
\end{align}
where the superscript $i \in \mathcal{I}_{\rm cal}$ indexes the instances as before and subscript $j \in [d]$ indexes the target dimensions. The resulting multivariate score can be written:
\begin{align} \label{eq:score_vector}
    S^{(i)} = [S^{(i)}_1, \dotsc, S^{(i)}_j] \in \mathbb{R}^d_+.
\end{align}
We then redefine the $\leq$ operator in \autoref{eq:conf_set} in terms of the vector partial ordering $\preccurlyeq$:
\begin{align} \label{eq:conf_set_joint}
{\Gamma}_{1-\alpha}(X^*) = \big \{Y \in \mathbb{R}^d: V(X^*, Y, \hat f) \preccurlyeq Q_{1-\alpha} \big \}, 
\end{align}
where $v \preccurlyeq w$ for $v, w \in \mathbb{R}^d$ if $v_1 \leq w_1 \dotsc, v_d \leq w_d$. To address the remaining question of modeling the CDF ${\hat F}$ of the score vectors in \autoref{eq:cdf_form}, we turn to copulas.

\paragraph{Copulas} \label{sec:vine_copulas} are statistical tools well suited for modeling the dependence structure of multivariate random variables.
\begin{Definition}[Copula]
    Given a random vector $S = [S_1,\dotsc, S_d]$, denote the marginal CDF for each component variable $S_j$, $j \in [d]$, as $F_j(s_j) = \mathbb{P}[S_j \leq s_j]$.
    The copula $C: [0, 1]^d \rightarrow [0, 1]$ of $S$ is the joint CDF of $U = [F_1(S_1), \dotsc, F_d(S_d)]$:
    \begin{align}
        C(u_1, \dotsc, u_d) = \mathbb{P}[F(S) \preccurlyeq u].
    \end{align}
\end{Definition}
Note that each $F_j(S_j)$ is the probability integral transform (PIT) of $S_j$ and follows a uniform distribution on the interval $[0, 1]$. Thus the copula function joins (couples) random variables, capturing their dependence structure, after all the complexities in their marginal distributions have been removed. The following theorem establishes its existence and uniqueness.
\begin{Theorem}[Sklar's theorem \citep{sklar1959fonctions}] \label{thm:sklar}
If random vector $S$ has a a joint CDF $F$ and marginal CDFs $F_1, \dotsc, F_d$, there exists a copula $C: [0, 1]^d \to [0, 1]$ such that $F(s_1, \dotsc, s_d) = C(F_1(s_1), \dotsc, F_d(s_d))$. The copula is unique if $F_1, \dotsc, F_d$ are continuous. The associated copula density is $f(u_1, \dotsc, u_d) = c(F_1(s_1), \dotsc, F_d(s_d))f_1(s_1) \cdots f_d(s_d)$, where $f_1, \dotsc, f_d$ are the marginal densities of $S$. 
\end{Theorem}

Estimation of $F$ therefore decomposes into estimating (1) the marginal distributions $F_1, \dotsc, F_d$ and (2) the copula function $C$. Kernel density estimation (KDE) and the empirical CDF enable nonparametric modeling of (1). For (2), we can choose a parametric copula family such as the Gaussian, Gumbel, and Clayton based on the observed correlation structure and estimate the parameters with maximum likelihood or moment matching. The parametric approach, however, is prone to model misspecification. 

The vine copula, or the pair copula construction, can accommodate dependencies of arbitrary complexity when a single standard copula model is inadequate. It is a hierarchical model composed of bivariate copula blocks organized in a sequence of nested trees, also called regular (R-) vines or simply vines \cite{joe1994multivariate}. The main idea is to factorize a $d$-dimensional copula density into a product of ${d(d-1)}/{2}$ bivariate conditional copula densities. Fitting a vine then boils down to choosing the bivariate copula models and the vine structure. The bivariate copulas can be estimated in a nonparametric manner \citep{chen2007nonparametric,Geenens2017,nagler2017nonparametric}. See \autoref{app:vine_copulas} for more details on nonparametric vine copula estimation. 

While nonparametric vine copulas enable flexible estimation of $F$, in the context of CP, we are only interested in its quantile functional, $F^{-1}(1-\alpha)$. Semiparametric methods help tailor our inference for this particular objective. 

\paragraph{Semiparametric inference} refers to the inference of nonparametric estimands, or functionals of the true data distribution $P$. Also called targeted or debiased learning, this approach has emerged as a major paradigm in recent years \citep{vanderlaan2011targeted}. If the target functional of interest is ``smooth'' (in ways we will make concrete below), it is sometimes possible to endow estimators with parametric properties like $\sqrt{n}$-consistency and asymptotic normality, even if our estimate of $P$ converges at a slower rate \citep{murphy2000profile,vanderlaan2011targeted,kennedy2022semiparametric,hines2022demystifying}. 

Suppose $P$ belongs to a set of probability measures $\mathcal{P}$ on a Polish sample space $(\mathcal{Z}, \mathcal{A})$. The target estimand is $\Psi(P)$, where $\Psi: \mathcal{P} \rightarrow \mathbb{R}$ is a measurable functional. We begin with the following key concept.  
\begin{Definition}[Efficient influence function]
    Consider the following mixture model,
    \begin{align} \label{eq:parametric_submodel}
        P_t = (1-t)P + t {\tilde P},
    \end{align}
    called the \emph{parametric submodel}, constructed by mixing $P$ and an arbitrary $\tilde P \in \mathcal{P}$ with parameter $t \in [0, 1]$. 
    The pathwise derivative of $\Psi$ at $P$ toward $\tilde P$ is:
    \begin{align}
        & \lim\limits_{t \to 0} \frac{\Psi(P_t) - \Psi(P)}{t} = \frac{d \Psi(P_t)}{dt} \bigg\rvert_{t=0}. \label{eq:gateaux}
    \end{align} 
    By the Riesz representation theorem, if this derivative exists, we can find a unique representer $\psi_P: \mathcal{Z} \rightarrow \mathbb{R}$ with finite variance under $P$ such that
    \begin{align} \label{eq:riesz}
        \frac{d \Psi(P_t)}{dt} \bigg\rvert_{t=0} &= \mathbb{E}_{ {\tilde P} - P} [\psi_P(Z)] =  \mathbb{E}_{ {\tilde P}} [\psi_P(Z)]. 
    \end{align}
    The second equality follows from restricting ourselves to mean-zero functions for $\psi_P$, as the middle expression is insensitive to constant shifts in $\psi_P$. When \autoref{eq:riesz} exists and is finite for any ``regular'' parametric submodel (for which the score $\tilde{P}(Z)/P(Z) - 1$ has finite variance), $\psi_P$ is called the \emph{efficient influence function} (EIF) and $\Psi$ is said to be pathwise differentiable \citep{van1991differentiable}. The EIF serves as a first-order distributional derivative that characterizes how sensitive $\Psi(P)$ is to changes in $P$ \citep{robins2017minimax,kennedy2022semiparametric}. 
\end{Definition}

\begin{figure}[t!]
\centering
\includegraphics[width=0.6\linewidth]{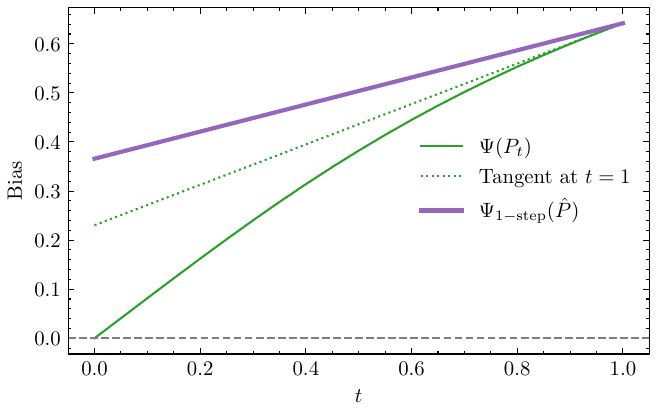}
\caption{\small One-step corrections visualized as linear projections. With the true target estimand being $\Psi(P)$, the green solid curve is the bias $\Psi(P_t) - \Psi(P)$, with $P_t$ defined in \autoref{eq:parametric_submodel}. The one-step estimator (purple solid line) can be viewed as approximating the slope of its tangent at $t=1$ (green dotted line) using the empirical distribution. \label{fig:bias_vs_epsilon}}
\end{figure}

Suppose we observe $n$ data points $Z^{(i)} \overset{\text{i.i.d.}}{\sim} P$ with $i \in [n]$. We may use the observations to obtain ${\hat P}$, an estimate of $P$. This induces the \emph{plug-in} estimator $\Psi({\hat P})$. Now let $\mathbb{P}_n$ denote the empirical distribution of the observations with associated density $\frac{1}{n} \sum_{i=1}^n \delta_{Z^{(i)}}$. The following procedure corrects $\Psi({\hat P})$ with an additive term based on the EIF.
\begin{Definition} The one-step estimator \citep{le1956asymptotic,pfanzagl1982existence,newey1998undersmoothing} is
    \begin{align}
        \Psi_{\rm 1\shorthyphen step}({\hat P}) = \Psi({\hat P}) + \mathbb{E}_{\mathbb{P}_n} [\psi_{\hat P}(Z^{(i)})]. 
    \end{align}
\end{Definition}
To see why this improves on the plug-in estimator, consider the distributional Taylor expansion,
\begin{align}
    \Psi(P) = \Psi({\hat P}) + \frac{d \Psi(P_t)}{dt} \bigg \rvert_{t=1} (0 - 1) + r_2(P,{\hat P}) \nonumber,
\end{align}
where $r_2$ is the remainder term \citep{yiu2023semiparametric}. Applying the Riesz representation theorem as in \autoref{eq:riesz} gives
$$\frac{d \Psi(P_t)}{dt} \bigg \rvert_{t=1} = -\mathbb{E}_P[ \psi_{\hat P}(Z)],$$
so we have 
\begin{align}
    \Psi(P) = \Psi({\hat P}) + \mathbb{E}_{P}[\psi_{\hat P}] + r_2(P,{\hat P}) \nonumber.
\end{align}
Denoting the estimation error of $\hat P$ as $\varepsilon$, we might expect $r_2 \sim \mathcal{O}(\varepsilon^2)$. This suggests  $\Psi({\hat P}) + \mathbb{E}_{P}[\psi_{\hat P}]$ would be better than the plug-in estimator $\Psi({\hat P})$. But $P$ is unknown, of course, so we take the expectation over $\mathbb{P}_n$. \autoref{fig:bias_vs_epsilon} helps interpret one-step correction as a linear projection \cite{fisher2021visually}. See \autoref{app:pathwise_diff} for an introduction to one-step estimation. 

\section{SEMIPARAMETRIC CP}

As described in \autoref{sec:problem_setup}, choosing an underlying prediction algorithm $\hat f: \mathcal{X} \to \mathbb{R}$ and a target-wise non-conformity score function $V(X_j, Y_j, {\hat f})$ induces score vectors in $\mathbb{R}^d$ evaluated on the calibration set, defined in \autoref{eq:score_targetwise} and \autoref{eq:score_vector}. We robustly estimate their $1-\alpha$ quantile by estimating the score distribution in a flexible non-parametric manner (\autoref{sec:density_estimation}) and applying the one-step correction to the plug-in quantile of the estimated distribution (\autoref{sec:semiparametric_correction}). See \autoref{alg:main} for the full algorithm.

\subsection{Nonparametric density estimation} \label{sec:density_estimation}

Motivated by recent developments in nonparametric estimation of copulas \citep{chen2007nonparametric}, we propose the use of nonparametric vine copulas described in \autoref{sec:background} to model the distribution $F$ of score vectors. First, we obtain the empirical marginal CDF:
\begin{align} \label{eq:marginal_ecdf}
    {\hat F}_j(s) = \frac{1}{n+1} \sum_{i \in \mathcal{I}_{\rm cal} \cup \{ \infty \}} \mathds{1}[S_j^{(i)} \leq s],
\end{align}
where $S_j^{(\infty)} \equiv \infty$ and again $n=|\mathcal{I}_{\rm cal}|$. Recall we index instances using superscripts $i \in [n]$ and index vector components using subscripts $j \in [d]$.  The resulting vector of uniform marginals can then be written
\begin{align} \label{eq:pit_ecdf}
    U^{(i)} \coloneqq [{\hat F}_1(S^{(i)}_1), \dotsc, {\hat F}_d(S^{(i)}_d)] \in [0, 1]^d.
\end{align}
For the bivariate copulas serving as building blocks for the vine, we opt for the transformation local likelihood kernel estimator (TLL), which allows KDE to be applied on the $[0, 1]^2$ constrained copula space \citep{geenens2014probit}. The resulting nonparametric vine copula ${\hat C}: [0, 1]^d \to [0, 1]$ yields the final estimated CDF
\begin{align} \label{eq:tll_copula}
    {\hat F}(s) = {\hat C}\left({\hat F}_1(s_1), \dotsc, {\hat F}_d(s_d)\right).
\end{align}
Parameters of the copula density, including the kernel bandwidth and vine structure, are fit using the Akaike information criterion (AIC), a consistent model selection criterion for nonparametric models \citep{akaike1974new}. The CDF can then be obtained by pairwise Monte-Carlo (MC) integration of the density. We use the \texttt{pyvinecopulib} library \citep{pyvinecopulib} for fitting and model selection.


\subsection{Semiparametric correction} \label{sec:semiparametric_correction}

Having obtained ${\hat F}$, we first compute the plug-in quantile:
\begin{align} \label{eq:plug_in_quantile}
    \Psi({\hat F}) = \{s \in \mathbb{R}^d: {\hat F}(s) \geq 1-\alpha \}.
\end{align}
This represents the $(1-\alpha)$-level curve of $\hat F$ illustrated in \autoref{fig:joint_quantile}. Note that the quantile is not unique; any quantile on the $(1-\alpha)$-level curve will do. The choice of which quantile to use for calibration depends on our preferences on the relative efficiency across the multiple targets. Suppose we would like a quantile with the smallest $L_1$ norm. We can obtain this point on the level curve by running optimization to find the quantile in the copula space and mapping it back to the score space via the inverse PIT:
\begin{align} 
    & U^* = \argmin_{U \in [0, 1]^d} ||U||_1 \ {\rm s.t.} \ {\hat C}(U) \geq 1- \alpha \label{eq:optim_copula} \\
    & \Psi({\hat F}) = Q_{1-\alpha}(U^*) = [{\hat F}_1^{-1}(U^*_1), \dotsc,  {\hat F}_d^{-1}(U^*_d)]. \label{eq:inverse_transform}
\end{align}
We use the CMA-ES algorithm \citep{hansen2003reducing} to optimize the copula in \autoref{eq:optim_copula}. 

We apply the one-step correction to $U^*$ rather than $\Psi(\hat F)$. The EIF for the generalized $1-\alpha$ quantile is
\begin{align} \label{eq:eif_quantile}
    \psi_{\hat C}(\tilde u) = \frac{(1-\alpha) - \mathds{1}[\tilde u \preccurlyeq U^*]}{|\nabla {\hat C}(U^*)|^2} \nabla {\hat C}(U^*),
\end{align}
where $\mathds{1}[b]$ is the indicator function that evaluates to 1 if $b$ is true and 0 otherwise. The derivation is provided in \autoref{app:eif_derivation}. The one-step estimator of $U^*$ is thus
\begin{align} \label{eq:one_step_quantile}
     U_{\rm 1\shorthyphen step} = U^* + \frac{1}{n}\sum_{i=1}^n \psi_{\hat C}(U^{(i)}).
\end{align}
We can then use \autoref{eq:inverse_transform} again to map $U_{\rm 1\shorthyphen step}$ to the score space and obtain $\Psi_{\rm 1\shorthyphen step}({\hat F})$.

\begin{figure}[t!]
\centering
\includegraphics[width=0.7\linewidth]{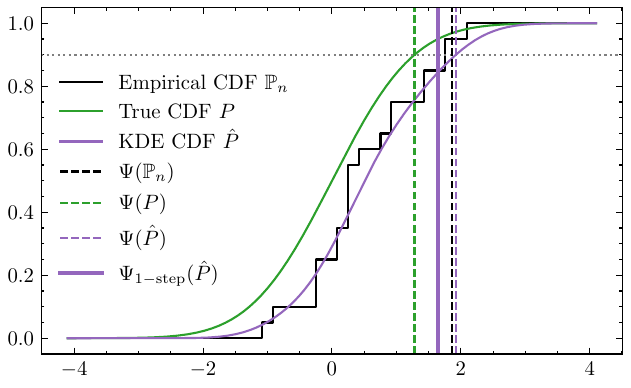}
\caption{\small Predicting the 0.9 quantile of a distribution based on 20 samples. Standard CP uses the empirical estimate (dashed black). Compared to the plug-in estimate from KDE (dashed purple), the one-step corrected estimate (solid purple) is closer to the true quantile (dashed green). \label{fig:cdf_quantiles}}  
\end{figure}

\autoref{fig:cdf_quantiles} illustrates one-step estimation in the single-target case where we seek to estimate the $1-\alpha=0.9$ quantile (green dashed line) of the true score distribution (green curve) based on $n{\rm =}20$ samples. Standard CP takes the empirical CDF (black curve) and extracts the $\lceil (n+1)(1-\alpha) \rceil/(n+1)$ quantile (dashed black line). Our approach is to ``augment'' the scores by fitting a nonparametric model for the CDF (purple curve). Here, the plug-in quantile (dashed purple line) lies very close to the empirical one and both overestimate the quantile, whereas the corrected estimate (solid purple line) offers a better approximation.

The final prediction set for our method is given in \autoref{eq:conf_set_joint}. As with \cite{messoudi2021copula}, it takes hyperrectangular shapes. For $V(X, Y, \hat f)_j = |Y_j - {\hat f}(X_j)|$, for example, the target-wise intervals are
\begin{align} 
    {\Gamma}_{1-\alpha}(X^*)_j = \{y \in \mathbb{R}: | y - {\hat f}(X^*)_j| \leq Q_{1-\alpha, j} \},
\end{align}
where $Q_{1-\alpha}$ can be $\Psi({\hat F})$ or $\Psi_{\rm 1\shorthyphen step}({\hat F})$.

\subsection{Theoretical results}

We informally state the theoretical guarantees of our method below. The coverage of \autoref{alg:main} is asymptotically exact by Theorem \autoref{thm:asymp_exact_cov} and is approximately valid by Theorem \autoref{thm:approx_valid}. Complete proofs can be found in \autoref{app:theory}. We assume the following:
\begin{itemize}
    \item The true distribution $F^*$ over the score vectors $S$ is absolutely continuous.
    \item The nonparametric copula estimator $\hat F$ is uniformly consistent (i.e. $\sup_{s\in\mathbb{R}^d} \Bigl|\hat{F}(s) - F^\ast(s)\Bigr| \overset{p}{\to} 0$ as $n\to\infty.$) under the standard smoothness and bandwidth conditions.
\end{itemize}

\begin{Theorem}[Asymptotic consistency] \label{thm:asymp_exact_cov}
Under the stated regularity conditions, prediction set $\Gamma_{1-\alpha}$ from \autoref{alg:main} constructed using the one-step correction $U^*$ satisfies
$$\mathbb{P}[Y^* \in \Gamma_{1-\alpha}(X^*) \to 1-\alpha \textrm{ as } n \to \infty,$$
for a test point $\left(X^*, Y^*\right)$. That is, its coverage is asymptotically exact.
\end{Theorem}
    The statement follows from the uniform consistency of the copula estimator $\hat F$, which implies consistency of its quantile, and asymptotic normality of the one-step estimator. 

\begin{Theorem}[Approximate validity] \label{thm:approx_valid}
Suppose that the total variation (TV) distance between the true distribution $F^*$ and the estimated distribution $\hat F$ is bounded by $\epsilon$. That is,
$$\sup_S | F^*(S) - {\hat F}(S) | \leq \epsilon.$$
Then the resulting prediction set $\Gamma_{1-\alpha}$ from \autoref{alg:main} constructed either from the plug-in quantile $U$ or the one-step corrected quantile $U^*$ is marginally valid at the $1-\alpha-\epsilon$ level:
$$\mathbb{P}[Y^* \in \Gamma_{1-\alpha}(X^*)] \geq 1-\alpha - \epsilon.$$
\end{Theorem}
The proof proceeds as follows. By the data processing inequality, the TV distance between $F^*$ and $\hat F$ upper-bounds the TV distance between their corresponding quantiles. The same holds between the quantile of $\hat F$ and the one-step corrected quantile, as the correction term does not add additional information.

\paragraph{Finite-sample validity.} In \autoref{app:split_alg}, we also introduce \autoref{alg:split}, a variant of \autoref{alg:main} that further splits the calibration set into $\mathcal{I}_{{\rm cal}, 1}$ and $\mathcal{I}_{{\rm cal}, 2}$. The set $\mathcal{I}_{{\rm cal}, 1}$ is used to build the empirical CDF $\hat F_j, \forall j \in [d]$ and the set $\mathcal{I}_{{\rm cal}, 2}$ with $n' = |\mathcal{I}_{{\rm cal}, 2}|$ is used to fit the empirical copula $\hat C$, from which the $\lceil(1-\alpha)(n'+1)/(n'+1)\rceil$ quantile is extracted. The extra split provides finite-sample validity by Theorem \ref{thm:split_finite_val}, because it makes the vector marginals $U$ between $\mathcal{I}_{{\rm cal}, 2}$ and the test set exchangeable.
 
\begin{algorithm}[t]
\caption{Semiparametric conformal prediction \label{alg:main}}
\begin{algorithmic}[1] 
\State {\bf Input:} Labeled data $\mathcal{I}$, test inputs $\mathcal{I}_{\rm test}$, target coverage level $1-\alpha$
\State {\bf Output:} Prediction sets ${\Gamma}_{1-\alpha}(X^{(i)})$ for $i \in \mathcal{I}_{\rm test}$
\State Split $\mathcal{I}$ into $\mathcal{I}_{\rm train}$ and $\mathcal{I}_{\rm cal}$
\State Train the underlying algorithm $\hat f$ on $\mathcal{I}_{\rm train}$ 
\For{$i \in \mathcal{I}_{\rm cal}$}
    \For{$j \in [d]$}
        \State $S^{(i)}_j \gets V(X^{(i)}_j, Y^{(i)}_j, {\hat f})$
    \EndFor
\EndFor
\For{$i \in \mathcal{I}_{\rm cal}$}
    \For{$j \in [d]$}
        \State ${\hat F}_j(s) \gets$ \autoref{eq:marginal_ecdf} \Comment{Marginal ECDF}
        \State $U^{(i)}_j \gets {\hat F}_j(S^{(i)}_j)$ \Comment{Uniform marginals}
    \EndFor
\EndFor
\State Fit the copula ${\hat C}$ on $\{U^{(i)}\}_{i \in \mathcal{I}_{\rm cal}}$
\State $U^* \gets$ \autoref{eq:optim_copula}
\State $Q_{1-\alpha} \gets Q_{1-\alpha}(U^*)$ from \autoref{eq:inverse_transform} 
\If{correction is True}
    \State $U^* \gets U_{\rm 1\shorthyphen step}$ from \autoref{eq:one_step_quantile} 
    \State $Q_{1-\alpha} \gets Q_{1-\alpha}(U^*)$ from \autoref{eq:inverse_transform}
\EndIf
\For{$i \in \mathcal{I}_{\rm test}$}
    \State $\Gamma_{1-\alpha}(X^{(i)}) \gets$ \autoref{eq:conf_set_joint}
\EndFor
\State \Return{$\Gamma_{1-\alpha}(X^{(i)}), \forall i \in \mathcal{I}_{\rm test}$}
\end{algorithmic} 
\end{algorithm}

\section{EXPERIMENTS}

\paragraph{Setup.}
We investigate split conformal calibration schemes for regression with $d$ real-valued targets. The underlying algorithm ${\hat f}$ fit on $\mathcal{I}_{\rm train}$ is a multi-task Lasso point predictor \citep{tibshirani1996regression}. For calibration schemes that operate on vector scores, including ours, we choose the $L_1$ non-conformity score in \autoref{eq:score_targetwise}. Scores are evaluated on the same calibration set of size $n=|\mathcal{I}_{\rm cal}|$. We fix $1-\alpha = 0.9$ and randomly split the data into training, calibration, and test sets. 

\paragraph{Baselines.} We compare calibration schemes that vary in the score definition and ${\hat F}$ estimation. \autoref{tab:calib_schemes} provides a detailed comparison of all the schemes.
\begin{tightemize}
    \item univariate calibration applied independently for each target at the $(1-\alpha)^{1/d}$ level (\textbf{independent})
    \item calibration applied to a \textbf{scalar score} defined as the $L_2$ norm\footnote{This score definition yields prediction sets shaped as $d$-dimensional balls. The $L_1$ norm would yield cross-polytopes ($d$-dimensional generalization of diamonds). The $L_\infty$ norm would yield hypercubes.} of the prediction error, $V(X, Y, \hat f) = ||Y - {\hat f}(X)||_2$ \citep{feldman2023calibrated}
    \item \textbf{empirical copula} fit on vector scores with the constraint $U^*_1 = \cdots = U^*_d$ \cite{messoudi2021copula}
    \item TLL R-vine fit on vector scores (\textbf{ours, plug-in})
    \item one-step correction applied to the above (\textbf{ours, corrected})
\end{tightemize}

\paragraph{Evaluation metrics.} We evaluate coverage and efficiency for each calibration scheme. Coverage is calculated as the empirical coverage on the test set $\mathcal{I}_{\rm test}$: 
\begin{align} \label{eq:coverage_metric}
    {\rm Coverage} = \frac{1}{|\mathcal{I}_{\rm test}|}\sum_{i \in \mathcal{I}_{\rm test}} \mathds{1}[Y^{(i)} \in {\Gamma}_{1-\alpha}(X^{(i)})].
\end{align}
The closer this value is to the target coverage rate $1-\alpha$, the better. Note that uninformatively large prediction sets will satisfy validity by overcovering (e.g., ${\Gamma}_{1-\alpha} = \mathbb{R}^d$ yields ${\rm Coverage} = 1 \geq 1- \alpha$). We prefer valid prediction sets that are \textit{efficient}, or have small sizes, so we also report their (hyper)volumes:
\begin{align}
    & {\rm Efficiency} = \frac{1}{|\mathcal{I}_{\rm test}|}\sum_{i \in \mathcal{I}_{\rm test}} \log {\rm Vol}\left({\Gamma}_{1-\alpha}(X^{(i)})\right).
\end{align}

\subsection{Synthetic example} \label{sec:synthetic}

We generate data using the {penicillin production} simulator from \citep{liang2021scalable}. This function takes in seven input parameters controlling penicillin production and outputs three real-valued targets ($d{\rm=}3$), namely penicillin yield, production time, and carbon dioxide emission. To simulate measurement errors, we add random Gaussian errors with scale equal to 10\% of the noiseless function output, so that $Y = f_0(X) + \varepsilon$ where $f_0$ is the test function and $\varepsilon \sim \mathcal{N}(0, (f_0(X)/10)^2)$. More details are in \autoref{app:synthetic_example}.

\begin{figure}[t!]
\centering
\includegraphics[width=0.5\linewidth]{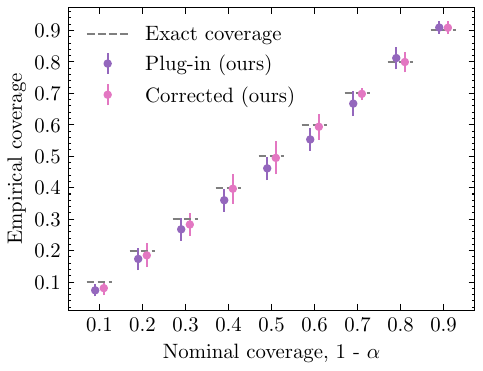}
\caption{\small One-step correction improves the coverage of the plug-in estimate at all target levels for the penicillin dataset. Error bars are stddev across ten random seeds. \label{fig:cov_vs_alpha}}
\end{figure}

\begin{figure}[t!]
\centering
\includegraphics[width=0.6\linewidth]{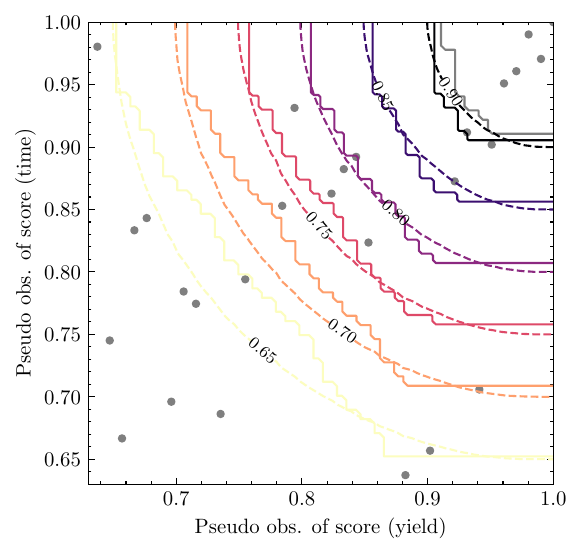}
\caption{\small Estimated (dashed) and true (solid) level curves at various quantile levels in the copula space. At the 0.9 level, the estimated curve (dashed black) falls between the true curve (solid black) and empirical (solid gray) curve computed from 96 points, some shown in gray dots. \label{fig:level_curves_penicillin2d}}
\end{figure}

As reported in \autoref{tab:synthetic_data_metrics}, our corrected method achieves coverage closest to the nominal level of $0.9$ and is the most efficient among the valid schemes. The one-step correction improves on the plug-in in terms of both coverage and efficiency. \autoref{fig:cov_vs_alpha} shows that the coverage improvement applies for all target levels. The level curves of the true, estimated, and empirical distributions for two penicillin targets are shown in \autoref{fig:level_curves_penicillin2d}.

On the other hand, the independent scheme is forced to extract an empirical marginal quantile at the $\lceil(0.9)^{1/d}(n+1)\rceil/(n+1) \approx 0.97$ level from only $n{\approx }100$ calibration instances, so become overconservative \citep{messoudi2021copula}. It cannot take advantage of the correlation in the residuals, particularly in the tails, that arise due to (tail) correlations in the target quantities, as visualized in \autoref{fig:penicillin_corner} and \autoref{fig:penicillin_scores_corner}. 

The scalar score suffers even more from inefficiency, because the $L_1$ score aggregates residuals from targets with drastically different distributions. In this example, the residuals for CO$_2$ emissions have a long upper tail relative to those for penicillin yield (see \autoref{fig:penicillin_corner}). Thus the scores are dominated by CO$_2$ residuals and the resulting quantile takes a large value. 

\autoref{app:calib_schemes} includes a complete set of experiments including our \autoref{alg:split} as well as other types of scalar scores ($L_2$ and $L_\infty$ norms) and a ``split'' scalar-score method that further splits the calibration set for normalizing the scores prior to scalarization.

\begin{table}[ht]
\caption{\small Metrics on the penicillin data ($d=3$, $n=96$). We report mean $\pm$ standard error across ten seeds. Target coverage is 0.9. For efficiency, lower is better. \label{tab:synthetic_data_metrics}}
\centering
\small
\begin{tabular}{lcc}
\toprule
\textbf{Method} & {\textbf{Coverage}} & {\textbf{Efficiency} $\downarrow$} \\ \midrule
Independent & $0.96 \pm 0.00$ & $4.1 \pm 0.1$ \\
Scalar score & $0.93 \pm 0.01$ & $4.5 \pm 0.1$ \\
Empirical copula & $0.93 \pm 0.01$ & $3.7 \pm 0.1$ \\
Plug-in (ours) & $0.91 \pm 0.01$ & $3.7 \pm 0.2$ \\
Corrected (ours) & $0.90 \pm 0.00$ & $3.5 \pm 0.1$ \\ 
Plug-in, split (ours) & $0.92 \pm 0.02$ & $3.6 \pm 0.3$ \\
Corrected, split (ours) & $0.90 \pm 0.03$ & $3.5 \pm 0.4$ \\
\bottomrule
\end{tabular}
\end{table}

\subsection{Real datasets} \label{sec:real_data}
\begin{table*}[ht]
\small
\centering
\caption{\small Metrics on the real data. We report mean $\pm$ standard error across five seeds. Target coverage is 0.9. For efficiency, lower is better. \label{tab:real_data_metrics}} 
\begin{tabular}{l|cc|cc|cc}
    \toprule
    \textbf{Method} & \multicolumn{2}{c|}{\textbf{Stock} ($d = 6, n = 63$)} & \multicolumn{2}{c|}{\textbf{Caco2+}  ($d = 6, n = 137$)} & \multicolumn{2}{c}{\textbf{rf1} ($d = 8, n = 225$)} \\
    & Coverage & Efficiency $\downarrow$ & Coverage & Efficiency $\downarrow$ & Coverage & Efficiency $\downarrow$ \\
    \midrule
    Independent     & $0.90 \pm 0.01$ & $-2.4 \pm 0.2$ & $0.95 \pm 0.01$ & $12.2 \pm 0.1$ & $0.96 \pm 0.01$ & $29.2 \pm 0.5$ \\
    Scalar score    & $0.90 \pm 0.01$ & $-1.8 \pm 0.3$ & $0.92 \pm 0.01$ & $28.3 \pm 0.1$ & $0.92 \pm 0.00$ & $27.7 \pm 0.3$ \\
    Empirical copula & $0.50 \pm 0.05$ & $-4.7 \pm 0.5$ & $0.42 \pm 0.08$ & $8.4 \pm 0.4$ & $0.42 \pm 0.12$ & $21.2 \pm 2.6$ \\
    Plug-in (ours)  & $0.87 \pm 0.02$ & $-2.9 \pm 0.2$ & $0.90 \pm 0.01$ & $11.0 \pm 0.1$ & $0.91 \pm 0.01$ & $25.0 \pm 0.2$ \\
    Corrected (ours) & $0.90 \pm 0.02$ & $-2.8 \pm 0.2$ & $0.93 \pm 0.01$ & $11.5 \pm 0.2$ & $0.91 \pm 0.01$ & $25.1 \pm 0.3$ \\
    \midrule
    \textbf{Method} & \multicolumn{2}{c|}{\textbf{rf2} ($d = 8, n = 225$)} & \multicolumn{2}{c|}{\textbf{scm1d} ($d = 16, n = 448$)} & \multicolumn{2}{c}{\textbf{scm20d} ($d = 16, n = 448$)} \\
    & Coverage & Efficiency $\downarrow$ & Coverage & Efficiency $\downarrow$ & Coverage & Efficiency $\downarrow$ \\
    \midrule
    Independent     & $0.96 \pm 0.01$ & $29.2 \pm 0.5$ & $0.96 \pm 0.01$ & $114.1 \pm 0.4$ & $0.96 \pm 0.01$ & $114.1 \pm 0.4$ \\
    Scalar score    & $0.92 \pm 0.01$ & $27.7 \pm 0.3$ & $0.89 \pm 0.01$ & $109.0 \pm 0.3$ & $0.89 \pm 0.01$ & $109.0 \pm 0.3$ \\
    Empirical copula & $0.42 \pm 0.12$ & $21.2 \pm 2.6$ & $0.75 \pm 0.05$ & $108.5 \pm 0.8$ & $0.75 \pm 0.05$ & $108.5 \pm 0.8$ \\
    Plug-in (ours)  & $0.90 \pm 0.01$ & $25.0 \pm 0.2$ & $0.92 \pm 0.01$ & $111.4 \pm 0.1$ & $0.92 \pm 0.01$ & $111.4 \pm 0.2$ \\
    Corrected (ours) & $0.91 \pm 0.01$ & $25.1 \pm 0.3$ & $0.91 \pm 0.01$ & $110.7 \pm 0.2$ & $0.90 \pm 0.01$ & $110.6 \pm 0.2$ \\
    \bottomrule
\end{tabular}
\vspace{-0.2cm}
\end{table*}

We report performance on a variety of real-world datasets with 6, 8, or 16 targets and a range of calibration-set sizes in \autoref{tab:real_data_metrics}. In all of these problems, our corrected scheme is valid and has empirical coverage closest to 0.9. It is also the most efficient among the valid schemes. For Stock \citep{liu2017using}, Caco2+ \citep{wang2016adme,park2023botied}, and rf2 \citep{spyromitros2016multi}, the correction is instrumental in making the plug-in estimate valid (above 0.9 coverage). At the same time, it does not significantly hurt efficiency.

With the other calibration schemes, we observe similar patterns as in the synthetic example. The independent method tends to overcover, and gives infinitely large prediction sets for the Stock dataset where we only have $n{\rm=}63$ instances from which to extract the empirical $\lceil(0.9)^{1/6} \times 64\rceil/64{\approx}0.98$ quantile. The scalar score overcovers even more and leads to very large prediction sets. The empirical copula exhibits high variance due to limited samples in the upper tails close to the target quantile. Its coverage falls substantially below the nominal level as well. 


\section{LIMITATIONS}

Just like the algorithm of \cite{messoudi2021copula}, our algorithm only guarantees approximate finite-sample validity. Loss of coverage can occur if the estimated copula $\hat C$ of the scores, constructed with $n$ data points, deviates significantly from the ``true'' copula in the $n \to \infty$ limit. The parametric copula families considered in \cite{messoudi2021copula} is prone to model misspecification bias; we aim at minimizing this using nonparametric vines. 

In our work, we used CMA-ES \citep{hansen2003reducing}, a gradient-free evolutionary algorithm, to optimize the copula. This step can be made more efficient, particularly for large $d$, by using gradient-based algorithms. Unlike empirical copulas and some classes of parametric copulas, which lend themselves to gradient descent algorithms \citep{sun2024copula}, nonparametric vines are more complicated for gradient computation. We leave this to future work.

\section{CONCLUSION}

We improve on conformal calibration for the multi-target regression setting. When there are many targets relative to the size of the calibration set, empirical quantile extraction suffers from curse of dimensionality and can exhibit high variance due to limited samples in the upper extreme of the non-conformity scores. Previous approaches to learn the joint score distribution and exploit its smoothness have been limited to parametric copula families, which are prone to model misspecification. Our method instead specifies a nonparametric model for full flexibility, but controls the plug-in bias on the quantile estimand of interest by way of the one-step semiparametric correction. We have demonstrated that our method yields prediction sets with desired coverage and competitive efficiency on a range of high-dimensional regression problems. Because it makes no assumptions about the form of the underlying algorithm or score function, it can be plugged in easily at the calibration step.

An interesting avenue for future work may be to specify the structure of the vine based on domain knowledge about the target dependence structure. We may use C-vines for when one target variable dominates or D-vines for temporal variables considered in \citep{sun2024copula}. 

\subsubsection*{Acknowledgements}
We thank Anastasios Angelopoulos for the insightful feedback.  


\bibliography{main}

\newpage
\onecolumn
\title{Semiparametric Conformal Prediction: \\
Supplementary Materials}
\part*{}
\section{NONPARAMETRIC VINE COPULAS} \label{app:vine_copulas}

This section describes the estimation and model selection procedure for nonparametric vine copulas. We start with a brief introduction to the concepts underlying copulas and vine copulas.

\subsection{Copulas} \label{app:copulas}
Let $S=[S_1, \dotsc, S_d]$ be a $d$-dimensional continuous random vector with joint distribution function $F$, joint density function $f$, marginal distribution functions $F_j$, and marginal density functions $f_j$ for $j \in [d]$. Sklar's theorem (Theorem \autoref{thm:sklar}) guarantees the existence of a copula $C: [0, 1]^d \rightarrow [0, 1]$ such that
\begin{align} \label{app:eq:joint_dist}
    F(s_1, \dotsc, s_d) = C(F_1(s_1), \dotsc, F_d(s_d)).
\end{align}
For absolutely continuous distributions, $C$ is unique. The marginals $u_j \coloneqq F_j(s_j), j \in [d]$, are uniformly distributed, being the result of the probability integral transform (PIT). This follows because
$$\mathbb{P}[F_j(S_j) \leq u_i] = \mathbb{P}[S_j \leq F_j^{-1}(u_j)] = F_j\left(F_j^{-1}(u_j) \right) = u_j, \quad \forall j \in [d].$$
Thus $C$ is a distribution function on the uniformly distributed marginals.

By partial differentiation of \autoref{app:eq:joint_dist}, we obtain the \textit{copula density} $c: [0, 1]^d \rightarrow [0, 1]$:
\begin{align} \label{app:eq:joint_copula_density}
    & f(s_1, \dotsc, s_d) = \frac{\partial^d F(s_1, \dotsc, s_d)}{\partial s_1 \cdots \partial s_d} = \underbrace{\frac{\partial^d C(u_1, \dotsc, u_d)}{\partial u_1 \cdots \partial u_d}}_{\eqqcolon c(u_1, \dotsc, u_d)} \prod_{i=1}^d \frac{d F_i(s_i)}{d s_i} = c(u_1, \dotsc, u_d) \prod_{i=1}^d f_i(s_i).
\end{align}
To model the joint distribution based on a dataset $\{(s_1^{(i)}, \dotsc, s_d^{(i)}) \}_{i=1}^n$, we can specify marginal models for each variable separately and join (couple) them with a copula model in a two-step process. In the first step, we apply the PIT to obtain the \textit{pseudo observations}, which are approximately uniform:
$$u^{(i)} = \left(F_1(s_1^{(i)}), \dotsc, F_d(s_d^{(i)}) \right), \quad \forall i \in [d].$$
Using parametric marginal models for $F_j, \forall j \in [d]$ amounts to an inference for margins (IFM) approach \citep{joe2005asymptotic}. In this work, we use the empirical marginal distribution \citep{genest1995semiparametric}.

In the second step, we specify the copula model and estimate its parameters by maximum likelihood estimation or moment matching. Parametric copula families such as the elliptical family (including the Gaussian and t-copulas) and Archimedean family (including the Gumbel, Clayton, and Frank copulas) can be restrictive in their assumptions about symmetry and tail dependence. Choosing a single parametric copula would imply that every pair of variables follows the same type of tail dependence, which may not be the case in practical applications. 

\subsection{Vine copulas}
Vine copulas are flexible models composed of bivariate building blocks. We will illustrate the pair copula decomposition for $d=3$, closely following the treatment in \cite{czado2022vine}. We can factorize the joint density as follows:
\begin{align}
f(s_1, s_2, s_3) = \redcolor{f_{3|12}(s_3 \mid s_1, s_2)} \bluecolor{f_{2|1}(s_2 \mid s_1)} f_1(s_1). \label{app:eq:joint_d3}
\end{align}
We use the notation $F_{j|k, \dotsc, l}$ and $f_{j|k, \dotsc, l}$ for the conditional distribution and density, respectively, of $S_j$ given $S_k, \dotsc, S_l$. We also use $c_{ij;k}(\cdot, \cdot;s_k)$ for the conditional copula density associated with the conditional density of $S_i, S_j$ given $S_k=s_k$, also called the \textit{pair copula}. From Theorem \autoref{thm:sklar}, we have
\begin{align}
\redcolor{f_{3|12}(s_3 \mid s_1, s_2)} = \frac{f_{13|2}(s_1, s_3 \mid s_2)}{f_{1|2}(s_1 \mid s_2)} = \frac{c_{13;2}(F_{1|2}(s_1 \mid s_2), F_{3|2}(s_3 \mid s_2); s_2) \cancel{f_{1|2}(s_1 \mid s_2)} \greencolor{f_{3|2}(s_3 \mid s_2)}
}{\cancel{f_{1|2}(s_1 \mid s_2)}}. \label{app:eq:f_3|12}
\end{align}

The conditional density $f_{2|1}$ and distribution $F_{2|1}$ can be written
\begin{align} 
    \bluecolor{f_{2|1}(s_2 \mid s_1)} &= c_{12}(F_1(s_1), F_2(s_2)) f_2(s_2), \label{app:eq:f_2|1} 
\end{align}
and, similarly, 
\begin{align}
    \greencolor{f_{3|2}(s_3 \mid s_2)} &= c_{23}(F_2(s_2), F_3(s_3)) f_3(s_3). \label{app:eq:f_3|2}
\end{align}

Substituting Equations \ref{app:eq:f_3|12}, \ref{app:eq:f_2|1}, and \ref{app:eq:f_3|2} into \autoref{app:eq:joint_d3}, we have
\begin{equation}
f(s_1, s_2, s_3) = \redcolor{c_{13;2}(F_{1|2}(s_1 \mid s_2), F_{3|2}(s_3 \mid s_2); s_2)} 
\ \redcolor{c_{23}(F_2(s_2), F_3(s_3))}
\ \bluecolor{c_{12}(F_1(s_1), F_2(s_2))} \redcolor{f_3(s_3)} \bluecolor{f_2(s_2)} f_1(s_1). \label{app:eq:pair_copula_decomposition}
\end{equation}

There exist two alternative decompositions of the joint density:
$$
f(s_1, s_2, s_3) = c_{23;1}(F_{2|1}(s_2 \mid s_1), F_{3|1}(s_3 \mid s_1); s_1) \ c_{13}(F_1(s_1), F_3(s_3))
\ c_{12}(F_1(s_1), F_2(s_2)) f_3(s_3) f_2(s_2) f_1(s_1),
$$
$$
f(s_1, s_2, s_3) = c_{12;3}(F_{1|3}(s_1 \mid s_3), F_{2|1}(s_2 \mid s_1); s_3) \ c_{13}(F_1(s_1), F_3(s_3))
\ c_{23}(F_2(s_2), F_3(s_3)) f_3(s_3) f_2(s_2) f_1(s_1).
$$
All three decompositions have pair copula terms of the form $c_{ij:k}(\cdot, \cdot ; s_k)$. Typically, we make the \textit{simplifying assumption} and ignore the explicit dependence on $s_k$ such that the arguments stand in for capturing the dependence. The simplified \textit{pair copula construction} (PCC), no longer an exact decomposition, for \autoref{app:eq:pair_copula_decomposition} is
\begin{align}
    f(s_1, s_2, s_3; \theta) &= c_{13;2}(F_{1|2}(s_1 \mid s_2), F_{3|2}(s_3 \mid s_2); \theta_{13;2}) \ c_{23}(F_2(s_2), F_3(s_3); \theta_{23}) \nonumber \\
& \ c_{12}(F_1(s_1), F_2(s_2), \theta_{12}) f_3(s_3) f_2(s_2) f_1(s_1), \label{app:eq:pair_copula_construction}
\end{align}
where $\theta = (\theta_{13;2}, \theta_{23}, \theta_{12})$ is the vector of parameters for the parametric copula densities $c_{13;2}, c_{23}$, and $c_{12}$. For estimating $\theta$ from a dataset $\{(s_1^{(i)},s_2^{(i)}, s_3^{(i)}) \}_{i=1}^n$, we can follow the two-step process outlined in \autoref{app:copulas}, first generating the pseudo observations $\{u^{(i)}\}_{i=1}^n = \{(u_1^{(i)}, u_2^{(i)}, u_3^{(i)}) \}_{i=1}^n$ and estimating $\theta$ via, for instance, maximum likelihood estimation:
$$\theta^* = {\rm argmax}_{\theta} \ \ell \left(\{u^{(i)}\}_{i=1}^n ; \theta \right) $$
$$\ell \left(\{u^{(i)}\}_{i=1}^n ; \theta \right) = \prod_{i=1}^{n} c_{13;2}(C_{1|2}(u^{(i)}_1 \mid u^{(i)}_2; \theta_{12}), C_{3|2}(u^{(i)}_1 \mid u^{(i)}_2; \theta_{23}); \theta_{13;2}) \\
\ c_{23}(u^{(i)}_2, u^{(i)}_3; \theta_{23}) \ c_{12}(u^{(i)}_1, u^{(i)}_2; \theta_{12}).$$

If parametric copula families are insufficient for fitting the pseudo observations, nonparametric approaches offer a more flexible fit. Transformation local likelihood kernel estimators (TLL) perform well \citep{geenens2014probit,nagler2017nonparametric}. We opt for the vine copula with TLL building blocks in this work.

The PCC described above generalizes to general $d$. The regular vine (R-vine) tree structure provides a framework for defining valid factorizations \citep{bedford2001probability,bedford2002vines}. A regular vine takes the form of a linked tree sequence, where the edges in one tree turn into the nodes of the next tree. Formally, a set of graphs $\mathcal{V} = (T_1, \dotsc, T_{d-1})$ is a regular vine on $d$ elements if it satisfies the following:
\begin{enumerate}
    \item $T_1$ is a tree with nodes $N_1 = \{1, \dotsc, d\}$ and edges $E_1$.
    \item For $j \geq 2$, $T_j$ is a tree with nodes $N_j = E_{j-1}$ and edges $E_j$.
    \item (Proximity condition) For $j \geq 2$ and $\{a, b\} \in E_j$, it holds that $|a \cap b | =1$. That is, if two nodes in $T_j$ have an edge between them, the corresponding edges in $T_{j-1}$ share a common node. 
\end{enumerate}

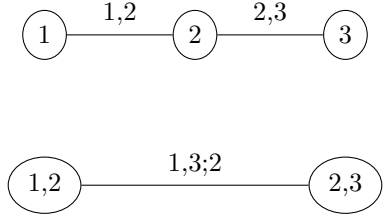
\begin{figure}[ht]
\centering
\begin{tikzpicture}
    \node[draw, ellipse] (1) at (0,0) {1};
    \node[draw, ellipse] (2) at (2,0) {2};
    \node[draw, ellipse] (3) at (4,0) {3};
    
    \draw (1) -- (2) node[midway, above] {1,2};
    \draw (2) -- (3) node[midway, above] {2,3};
    
    \node[draw, ellipse] (12) at (0,-2) {1,2};
    \node[draw, ellipse] (23) at (4,-2) {2,3};
    
    \draw (12) -- (23) node[midway, above] {1,3;2};
\end{tikzpicture}
\caption{Graphical diagram of the vine structure corresponding to the PCC in \autoref{app:eq:pair_copula_construction}. \label{fig:graphical_pcc}}
\end{figure}

Each edge is associated with a pair copula. The example in \autoref{app:eq:pair_copula_construction} has the vine structure $N_1 = \{1, 2, 3\}, E_1 = \{\{1, 2\}, \{2, 3\}\}$, $N_2=\{\{1, 2\}, \{2, 3\}\}$, and $E_2=\{1, 3;2\}$, for which \autoref{fig:graphical_pcc} provides a graphical representation. To simplify the notation for the node and edge sets, let us define the complete union associated with a given edge $e \in E_k$:
$$A_e \coloneqq \{k \in N_1 \mid \exists e_1 \in E_1, \dotsc, e_{k-1} \in E_{k-1} \textrm{ s.t. } j \in e_1 \in \dotsc \in e_{k-1} \in e\}.$$
An edge $e = \{a, b\}$ corresponds to the pair copula $c_{a, b ;D_e}$, for which the \textit{conditioning set} $D_e$ of $e$ is $D_e = A_a \cap A_b$ and the \textit{conditioned sets} $C_{e, a}, C_{e, b}$ are $C_{e, a} = A_a \backslash D_e, \ C_{e, b} = A_b \backslash D_e$. If $a=\{1, 2\}$ and $b=\{2, 3\}$ for an edge $e = \{a, b\}$, for instance, then $A_a=\{1, 2\}$, $A_b = \{2, 3\}$, and $D_e = \{2\}$. Then the conditioned sets are $C_{e, a} = \{1\}$ and $C_{e, b} = \{3\}$. 

In summary, one can model the $d$-variate joint distribution for a random vector $S = (S_1, \dotsc, S_d)$ with a regular vine distribution specified by the tuple $(\mathcal{F}, \mathcal{V}, \mathcal{B})$, where the $\mathcal{F} = (F_1, \dotsc, F_d)$ is the tuple of the marginal distribution functions, $\mathcal{V}$ is the vine structure, and $\mathcal{B} = \{C_e| e \in \mathcal{E_j}; j \in [d-1] \}$, the set of bivariate copulas $C_e$ associated with the conditional distribution of $S_{C_{e, a}}$ and $S_{C_{e, a}}$ given $S_{D_e}$. 
\begin{Theorem}[Existence of a regular vine distribution \citep{bedford2002vines}] \label{thm:existence_vine}
Given the tuple $(\mathcal{F}, \mathcal{V}, \mathcal{B})$, then there exists a valid $d$-dimensional distribution $F$ with density
$$
f_{1,\dots,d}(s_1, \dots, s_d) = f_1(s_1) \times \cdots \times f_d(s_d) \\
\times \prod_{i=1}^{d-1} \prod_{e \in {E}_i} c_{c_{e,a}, c_{e,b}; D_e}\left(F_{c_{e,a}|D_e}(s_{c_{e,a}} \mid {s}_{D_e}), F_{c_{e,b}|D_e}(s_{c_{e,b}} \mid {s}_{D_e})\right),
$$
such that
$$
F_{c_{e,a}, c_{e,b}|D_e}(s_{c_{e,a}}, s_{c_{e,b}} \mid {s}_{D_e}) = C_e \left( F_{c_{e,a}|D_e}(s_{c_{e,a}} \mid {s}_{D_e}), F_{c_{e,b}|D_e}(s_{c_{e,b}} \mid {s}_{D_e}) \right).
$$
\end{Theorem}
The parameters of the bivariate copulas can be estimated via maximum likelihood. While the number of valid vine structures grows super-exponentially with $d$, as $d! \times 2^{(d-2)(d-3)/2 - 1}$ \citep{napoles2010number}, the greedy selection algorithm by \cite{dissmann2013selecting} enables efficient exploration.



\section{EFFICIENT INFLUENCE FUNCTION} \label{app:eif_derivation}

In this section, we provide derivations of the efficient influence function (EIF) of quantile estimands. We follow the \emph{point-mass contamination} method. It works by perturbing the estimand in the direction of a point mass at a single observation ${\tilde s}$ to obtain the mixture parametric submodel in \autoref{eq:parametric_submodel} with ${\tilde P} = \delta_{\tilde s}$:
\begin{align} \label{eq:point_mass_contamination}
    P_t = (1-t) P + t \delta_{\tilde s}.
\end{align}

We then take
\begin{align} \label{eq:app:eif_derivative}
    \psi_P({\tilde s}) = \frac{d \Psi(P_t)}{dt} \bigg |_{t=0},
\end{align}
also called the G{\^a}teaux derivative. 

\subsection{Univariate quantile} \label{sec:eif_1d}

The derivation here closely follows the presentation in \cite{hines2022demystifying}. The estimand of interest is $\Psi_{1-\alpha}(P) = F^{-1}(1-\alpha)$. The quantile can be implicitly written 
\begin{align*}
    \int_{-\infty}^{\Psi_{1-\alpha}(P)} f(s) ds = 1-\alpha,
\end{align*}
where $f$ is the density function. 

In terms of the parametric submodel, we have
\begin{align*}
    \int_{-\infty}^{\Psi_{1-\alpha}(P_t)} f_t(s) ds = 1-\alpha,
\end{align*}
with $f_t(s) = (1-t) f(s) + t\delta_{\tilde s}(s)$.

Differentiating both sides with respect to $t$ using the Leibniz integral rule,
\begin{align*}
    f_t\left( \Psi_{1-\alpha} (P_t) \right)\frac{d \Psi_{1-\alpha}(P_t)}{dt} + \int_{-\infty}^{\Psi_{1-\alpha}(P_t)} \frac{d f_t(s)}{dt} ds = 0.
\end{align*}

With $F_t(s) = (1-t) F(s) + t\mathds{1}[{\tilde s} \leq s]$,
\begin{align*}
    \frac{d \Psi_{1-\alpha}(P_t)}{dt} &= \frac{-1}{f_t\left( \Psi_{1-\alpha} (P_t) \right)}  \int_{-\infty}^{\Psi_{1-\alpha}(P_t)} \frac{d f_t(s)}{dt} ds \\
    &= \frac{1}{f_t\left( \Psi_{1-\alpha} (P_t) \right)} \int_{-\infty}^{\Psi_{1-\alpha}(P_t)} \left( f(s) - \delta_{\tilde s}(s)  \right) ds \\
    &= \frac{ F(F_t^{-1}(1-\alpha) ) - \mathds{1}[\tilde s \leq \Psi_{1-\alpha}(P_t)] }{f_t\left( \Psi_{1-\alpha} (P_t) \right)}.
\end{align*}

Evaluating at $t=0$, we then have
\begin{align*}
    \psi_P({\tilde s}) &= \frac{ 1-\alpha - \mathds{1}[\tilde s \leq \Psi_{1-\alpha}(P)] }{f\left( \Psi_{1-\alpha} (P) \right)} \\
    &= \frac{\mathds{1}[\Psi_{1-\alpha}(P) \leq {\tilde s}] - \alpha}{f\left( \Psi_{1-\alpha} (P) \right)}. \qed
\end{align*}

The pinball loss used for quantile regression is $l_{1-\alpha}(x) = x ( \mathds{1}[x\geq 0] -\alpha)$ \citep{koenker1978regression}. In the numerator of the last expression, one may recognize its derivative $l'_{1-\alpha}(x) =  \mathds{1}[x\geq 0] -\alpha$, namely $l'_{1-\alpha}({\tilde s} - \Psi_{1-\alpha} (P))$.

\subsection{Multivariate quantile} \label{app:eif_multivariate}

For $d > 1$, the parametric submodel is $F_t(s) = (1-t) F(s) + t\mathds{1}[{\tilde s}_1 \leq s_1, \dotsc, {\tilde s}_d \leq s_d]$, with $s = [s_1, \dotsc, s_d]$. We proceed similarly as before, but directly in terms of the CDF to avoid dealing with multiple integrals. To simplify notation, let us first define $\Psi_{1-\alpha}(P_t) \coloneqq g(t)$ and $\Psi_{1-\alpha}(P) \coloneqq g$. 

The quantile under the parametric submodel satisfies
\begin{align*}
    (1-t) F(g(t)) + t \mathds{1}[\underbrace{{\tilde s}_1 \leq g(t)_1, \dotsc, {\tilde s}_d \leq g(t)_d}_{\tilde s \preccurlyeq g(t)}] = 1-\alpha.
\end{align*}

Differentiating both sides with respect to $t$ and setting $t=0$,
\begin{align}
    & -F(g) + \frac{\partial F}{\partial s_1} \bigg |_{s=g} \frac{d g_1}{d t} + \cdots + \frac{\partial F}{\partial s_d} \bigg |_{s=g} \frac{d g_d}{d t} + \mathds{1}[\tilde s \preccurlyeq g] = 0 \nonumber \\ 
    & \implies -(1-\alpha) + \frac{\partial F}{\partial s_1} \bigg |_{s=g} \frac{d g_1}{d t} + \cdots + \frac{\partial F}{\partial s_d} \bigg |_{s=g} \frac{d g_d}{d t} + \mathds{1}[\tilde s \preccurlyeq g] = 0 \nonumber \\ 
    & \implies \frac{\partial F}{\partial s_1} \bigg |_{s=g} \frac{d g_1}{d t} + \cdots + \frac{\partial F}{\partial s_d} \bigg |_{s=g} \frac{d g_d}{d t} = (1-\alpha) - \mathds{1}[\tilde s \preccurlyeq g] \label{eq:app:total_diff}.
\end{align}

Recall the EIF is 
\begin{align*}
    \psi_P(\tilde s) = \left(\frac{d g_1}{d t}, \dotsc, \frac{d g_d}{d t} \right).
\end{align*}

We define the gradient $\nabla F = \left(\frac{\partial F}{\partial s_1}, \dotsc, \frac{\partial F}{\partial s_d} \right)$ and assume that $\psi_P(\tilde s)$ is proportional to $\nabla F(g)$:
\begin{align} \label{eq:app:prop_assumption}
    \psi_P(\tilde s) = \lambda \nabla F(g).
\end{align}

We can then rewrite \autoref{eq:app:total_diff} as
\begin{align}
    & \nabla F(g) \cdot (\lambda \nabla F(g)) = (1-\alpha) - \mathds{1}[\tilde s \preccurlyeq g] \nonumber \\
    & \implies \lambda | \nabla F(g)|^2 = (1-\alpha) - \mathds{1}[\tilde s \preccurlyeq g] \nonumber \\
    & \implies \lambda = \frac{(1-\alpha) - \mathds{1}[\tilde s \preccurlyeq g]}{| \nabla F(g)|^2} \label{eq:app:lambda_isolated},
\end{align}
where $|\cdot|^2$ is the $L_2$ norm. Thus \autoref{eq:app:prop_assumption} and \autoref{eq:app:lambda_isolated} together give
\begin{align*}
    \psi_P(\tilde s) = \frac{(1-\alpha) - \mathds{1}[\tilde s \preccurlyeq g]}{|\nabla F(g)|^2} \nabla F(g). \qed
\end{align*}

Note that, for $d=1$, this reduces to 
\begin{align*}
    \psi_P(\tilde s) = \frac{(1-\alpha) - \mathds{1}[\tilde s \leq g]}{f(g)^2} f(g) =  \frac{(1-\alpha) - \mathds{1}[\tilde s \leq g]}{f(g)} = \frac{(1-\alpha) - (1 - \mathds{1}[g \leq \tilde s])}{f(g)} = \frac{\mathds{1}[g \leq \tilde s] - \alpha}{f(g)},
\end{align*}
recovering the result from \autoref{sec:eif_1d}. 

\section{PATHWISE DIFFERENTIABLE ESTIMANDS} \label{app:pathwise_diff}

This section provides an introduction to one-step estimation in the context of shrinking the plug-in bias. Based on a decomposition of the plug-in bias, we will show that nonparametric estimation followed by one-step correction helps us control for the leading terms contributing to the bias. Once again, denote the true data-generating distribution as $P$ and our estimand of interest as $\Psi(P)$. Suppose we have obtained $\hat P$, our data-adaptive estimate of $P$. Define the parametric submodel:
\begin{align*}
    P_t = (1-t) P + t {\hat P}.
\end{align*}

The distributional Taylor expansion, also called the von Mises expansion, of $\Psi(P)$ about $t=1$ can be written:
\begin{align} \label{app:eq:von_mises_expansion}
    \Psi(P) = \Psi({\hat P}) + \frac{d \Psi(P_t)}{dt} \bigg \rvert_{t=1} (0 - 1) + r_2(P,{\hat P}) =
    \Psi({\hat P}) + \mathbb{E}_{P}[\psi_{\hat P}] + r_2(P,{\hat P}),
\end{align}
where $r_2$ is the remainder term. In what follows, we will use the notation $\mathbb{E}_{P}[\cdot] = P[\cdot]$. Let us now consider the scaled plug-in bias:
\begin{align} \label{app:eq:scaled_bias}
    & \sqrt{n} \left( \Psi({\hat P}) - \Psi(P) \right) =  -\sqrt{n}P[\psi_{\hat P}] - \sqrt{n}r_2(P,{\hat P}) \\
    &= \sqrt{n}\left(\mathbb{P}_n - P \right)[\psi_P] + \sqrt{n}{\mathbb{P}_n}[\psi_{\hat P}] - \sqrt{n}\left(\mathbb{P}_n - P\right)[\psi_P] - \sqrt{n}{\mathbb{P}_n}[\psi_{\hat P}] -\sqrt{n}P[\psi_{\hat P}] - \sqrt{n}r_2(P,{\hat P}) \\
    & = \sqrt{n}\left(\mathbb{P}_n - P \right)[\psi_P] + \sqrt{n}\left(\mathbb{P}_n - P \right)[\psi_{\hat P} - \psi_P] -\sqrt{n}P[\psi_{\hat P}] - \sqrt{n}r_2(P,{\hat P}) \\
    &= \underbrace{\sqrt{n}\mathbb{P}_n[\psi_P]}_{({\rm a})}  - \underbrace{\sqrt{n}P[\psi_P]}_{(\rm b)} + \underbrace{\sqrt{n}\left(\mathbb{P}_n - P \right)[\psi_{\hat P} - \psi_P]}_{(\rm c)} -\underbrace{\sqrt{n}P[\psi_{\hat P}]}_{(\rm d)} - \underbrace{\sqrt{n}r_2(P,{\hat P})}_{(\rm e)}.
\end{align}
Term $(\rm b) = 0$ as the influence function satisfies $P[\psi_P]= 0$, and term $(\rm a)$ asymptotically converges to a normal around zero by the central limit theorem. As discussed in \citep{hines2022demystifying}, the term $(\rm c)$ and $(\rm e)$ can be shown to converge to zero under certain conditions as well. 

The drift term $(\rm d)$, however, persists and may even diverge in some cases. Notably, it would not even converge to zero if $\psi_{P}$ were used in place of $\psi_{\hat P}$. We would like to remove this term,
\begin{align*}
    ({\rm d}) = \sqrt{n}P[\psi_{\hat P}] \approx \sqrt{n}{\mathbb{P}_n}[\psi_{\hat P}] = \frac{1}{\sqrt{n}} \sum_{i=1}^n \psi_{\hat P}(S^{(i)}).
\end{align*}
The one-step estimator simply adjusts the plug-in estimator such that the bias due to this term is zero:
\begin{align}
    \Psi_{\rm 1\shorthyphen step}({\hat P}) = \Psi(\hat P) + \frac{1}{n} \sum_{i=1}^n \psi_{\hat P}(S^{(i)}).
\end{align}

It is difficult to reason about the asymptotic behavior of term $(\rm c)$, the empirical process term, because the data-adaptive estimate $\hat P$ is involved. For the sake of analysis, let us assume that $\hat P$ is fit on an independent dataset from the dataset over which the expectation of EIF is taken, so that we can condition on $\hat P$. By Chebyshev's inequality, then,
\begin{align}
    P[\left( \psi_{\hat P} - \psi_P \right)^2 | \hat P] = o_p(1) \implies ({\rm c}) \coloneqq \sqrt{n}\left(\mathbb{P}_n - P \right)[\psi_{\hat P} - \psi_P] = o_p(1).
\end{align}
In turn, a sufficient condition for $P[\left( \psi_{\hat P} - \psi_P \right)^2 | \hat P] = o_p(1)$ is that $\hat P$ converges to $P$ in probability.  Withholding an independent subset of the data for fitting $\hat P$ thus helps shrink term (c). See \citep{vanderlaan2011targeted} and \citep{chernozhukov2018double} for analyses of K-fold cross validation. Empirically, we find that the effect of term $(\rm c)$ is small, and choose to use as much of the data for one-time copula fitting as possible rather than use K-fold splits (which would increase the computational overhead of our method) or holding out a split (which would hurt the copula fit). 

\section{SPLIT ALGORITHM}
\label{app:split_alg}

We present \autoref{alg:split}, a variant of \autoref{alg:main} that uses an extra split of the calibration set. Differences from \autoref{alg:main} are indicated in orange. One split of the calibration set is used to compute the marginal ECDFs and the other to fit the copula. This variant guarantees finite-sample distribution-free validity by Theorem \autoref{thm:split_finite_val}.

\begin{algorithm}[H]
\caption{Semiparametric conformal prediction, split version \label{alg:split}}
\begin{algorithmic}[1] 
\State {\bf Input:} Labeled data $\mathcal{I}$, test inputs $\mathcal{I}_{\rm test}$, target coverage level $1-\alpha$
\State {\bf Output:} Prediction sets ${\Gamma}_{1-\alpha}(X^{(i)})$ for $i \in \mathcal{I}_{\rm test}$
\State Split $\mathcal{I}$ into $\mathcal{I}_{\rm train}$, \textcolor{orange}{$\mathcal{I}_{\rm cal, 1}$, and $\mathcal{I}_{\rm cal, 2}$}
\State Train the underlying algorithm $\hat f$ on $\mathcal{I}_{\rm train}$ 
\For{$i \in \mathcal{I}_{\rm cal}$}
    \For{$j \in [d]$}
        \State $S^{(i)}_j \gets V(X^{(i)}_j, Y^{(i)}_j, {\hat f})$
    \EndFor
\EndFor
\For{ \textcolor{orange}{$i \in \mathcal{I}_{\rm cal, 1}$} }
    \For{$j \in [d]$}
        \State ${\hat F}_j(s) \gets$ \autoref{eq:marginal_ecdf} \Comment{Marginal ECDF}
        \State $U^{(i)}_j \gets {\hat F}_j(S^{(i)}_j)$ \Comment{Uniform marginals}
    \EndFor
\EndFor
\State{ \textcolor{orange}{Fit the copula ${\hat C}$ on $\{U^{(i)}\}_{i \in \mathcal{I}_{\rm cal, 2}}$ } }
\State $U^* \gets$ \autoref{eq:optim_copula}
\State $Q_{1-\alpha} \gets Q_{1-\alpha}(U^*)$ from \autoref{eq:inverse_transform} 
\If{correction is True}
    \State $U^* \gets U_{\rm 1\shorthyphen step}$ from \autoref{eq:one_step_quantile} 
    \State $Q_{1-\alpha} \gets Q_{1-\alpha}(U^*)$ from \autoref{eq:inverse_transform}
\EndIf
\For{$i \in \mathcal{I}_{\rm test}$}
    \State $\Gamma_{1-\alpha}(X^{(i)}) \gets$ \autoref{eq:conf_set_joint}
\EndFor
\State \Return{$\Gamma_{1-\alpha}(X^{(i)}), \forall i \in \mathcal{I}_{\rm test}$}
\end{algorithmic} 
\end{algorithm}

\section{THEORETICAL RESULTS}
\label{app:theory}

\noindent\textbf{Theorem \ref{thm:asymp_exact_cov}.}
\begin{proof}
Define the $(1-\alpha)$-quantile of $F^\ast$ as
$$
Q_{1-\alpha} = \inf_s \{||s||_1: F^\ast(s) = 1-\alpha \},
$$
and let $\hat{Q}_{1-\alpha}$ be the plug-in estimator obtained from $\hat{F}$. Since the quantile functional $\Psi: F\mapsto Q_{1-\alpha}$ is Hadamard differentiable under the stated regularity conditions, we have
$$
\sqrt{n}\Bigl(\hat{Q}_{1-\alpha} - Q_{1-\alpha}\Bigr) = \frac{1}{\sqrt{n}}\sum_{i=1}^n \psi(S^{(i)}) + o_p(1),
$$
where $\psi: \mathbb{R}^d\to\mathbb{R}^d$ is the EIF for the quantile functional. The one-step corrected estimator is defined by
$$
\hat{Q}_{1-\alpha}^{\text{1-step}} = \hat{Q}_{1-\alpha} + \frac{1}{n}\sum_{i=1}^n \psi(S^{(i)}).
$$
It then follows that
$$
\sqrt{n}\Bigl(\hat{Q}_{1-\alpha}^{\text{1-step}} - Q_{1-\alpha}\Bigr) = o_p(1),
$$
so that $\hat{Q}_{1-\alpha}^{\text{1-step}}$ is a $\sqrt{n}$-consistent estimator of $Q_{1-\alpha}$.

Now consider a test point $(X^\ast,Y^\ast)$ with associated nonconformity score $S^\ast = V(X^\ast,Y^\ast,\hat{f}).$
Since the prediction set is defined by
$$
\Gamma_{1-\alpha}(X^\ast)=\{Y \in\mathbb{R}^d: V(X^\ast, Y, \hat{f}) \preceq \hat{Q}_{1-\alpha}^{\text{1-step}}\},
$$
we have
$$
\mathbb{P}\bigl(Y^\ast\in\Gamma_{1-\alpha}(X^\ast)\bigr) = \mathbb{P}\bigl(S^\ast\preceq \hat{Q}_{1-\alpha}^{\text{1-step}}\bigr).
$$
By the definition of $\hat{F}$ and using the uniform consistency of $\hat{F}$ together with the continuity of $F^\ast$, it holds that
$$
\hat{F}\bigl(\hat{Q}_{1-\alpha}^{\text{1-step}}\bigr) \overset{p}{\to} F^\ast\bigl(Q_{1-\alpha}\bigr) = 1-\alpha.
$$
Thus, as $n\to\infty$,
$$
\mathbb{P}\bigl(S^\ast\preceq \hat{Q}_{1-\alpha}^{\text{1-step}}\bigr) = \hat{F}\bigl(\hat{Q}_{1-\alpha}^{\text{1-step}}\bigr) \to 1-\alpha.
$$

\end{proof}

\noindent\textbf{Theorem \ref{thm:approx_valid}.} 
\begin{proof}
We assumed that
$$
\sup_{S} \Bigl| F^*(S) - \hat{F}(S) \Bigr| \le \epsilon.
$$
Let ${\hat Q}_{1-\alpha}$ be defined as the $(1-\alpha)$-quantile of $\hat{F}$, i.e.,
$$
\hat{F}({\hat Q}_{1-\alpha}) = 1-\alpha.
$$
Recall that the prediction set using the plug-in quantile is defined as
$$\Gamma_{1-\alpha}(X^*) = \{Y \in \mathbb{R}^d : V(X^*, Y, \hat{f}) \preccurlyeq {\hat Q}_{1-\alpha}\}.$$

Note that the mapping from the distribution $F$ to the quantile is Lipschitz continuous under the TV distance. More precisely, by the data processing inequality, which states that for any measurable function $g$, 
$$
d_{\mathrm{TV}}\bigl(P\circ g^{-1}, Q\circ g^{-1}\bigr) \le d_{\mathrm{TV}}(P,Q),
$$
we have for the quantile
$$
A(F') = \inf_s \{||s||_1: F' = 1-\alpha \},
$$
that
$$
\Bigl| A(F) - A(\hat F) \Bigr| \le \sup_{S} \Bigl| F^*(S) - \hat{F}(S) \Bigr| \le \epsilon.
$$

Since $\hat{F}({\hat Q}_{1-\alpha}) = 1-\alpha$, it follows that
$$
F^*({\hat Q}_{1-\alpha}) \geq 1-\alpha - \epsilon.
$$
Now the coverage is
$$
\mathbb{P}\Bigl[ Y^* \in \Gamma_{1-\alpha}(X^*) \Bigr] 
= \mathbb{P}\Bigl[ V(X^*,Y^*,\hat{f}) \le {\hat Q}_{1-\alpha} \Bigr]
= F^*({\hat Q}_{1-\alpha}) \geq 1-\alpha - \epsilon.
$$ as desired. If instead we use the one-step corrected quantile to form the prediction set $\Gamma_{1-\alpha}^{\text{1-step}}$, then by design the correction term does not introduce additional error beyond the estimation error of $\hat{F}$. Hence, the same argument applies, and we again obtain
$$
\mathbb{P}\Bigl[ Y^* \in \Gamma_{1-\alpha}^{\text{1-step}}(X^*) \Bigr] \ge 1-\alpha-\epsilon.
$$
\end{proof}

\begin{Theorem}[Finite-sample validity of \autoref{alg:split}]
\label{thm:split_finite_val}
Assume that the full dataset (including calibration data and the test point) is exchangeable, and that the calibration set is split into two disjoint subsets, 
$$
I_{\mathrm{cal}} = I_{\mathrm{cal},1} \cup I_{\mathrm{cal},2},
$$
where the set $I_{\mathrm{cal},1}$ is used to construct the marginal empirical CDFs and to fit the copula, and the set $I_{\mathrm{cal},2}$ (with $n' \equiv |I_{\mathrm{cal},2}|$) is used to construct the empirical copula, from which the $\lceil(1-\alpha)(n'+1) \rceil/(n'+1)$-quantile is extracted. Then the prediction set 
$$
\Gamma_{1-\alpha}(X^*) = \{y\in \mathbb{R}^d: V(X^*, y, \hat{f}) \preceq Q_{1-\alpha}\},
$$
constructed with the quantile $Q_{1-\alpha}$ from $I_{\mathrm{cal},2}$ satisfies
$$
\mathbb{P}\Bigl( Y^* \in \Gamma_{1-\alpha}(X^*) \Bigr) \ge 1-\alpha.
$$
\end{Theorem}

\begin{proof}
Since the data are exchangeable, the procedure of splitting the calibration set into $I_{\mathrm{cal},1}$ and $I_{\mathrm{cal},2}$ guarantees that the scores computed on $I_{\mathrm{cal},2}$ are independent of the marginal and copula estimates that were obtained from $I_{\mathrm{cal},1}$. Denote by
$$
S_1, S_2, \ldots, S_{n'} 
$$
the scores computed on the samples in $I_{\mathrm{cal},2}$, and let $S^*$ be the score computed on the test point $(X^*,Y^*)$. By construction, the scores $S_1, \ldots, S_{n'}$ and $S^*$ are exchangeable.

Let $Q_{1-\alpha}$ be the $\lceil(1-\alpha) (n'+1)\rceil$-th order statistic of the $n'$ calibration scores. By the standard conformal prediction argument (see, e.g., \cite{vovk2005algorithmic, lei2014distribution}), the rank of $S^*$ among the $n'+1$ scores is uniformly distributed. We thus have
$$
\mathbb{P}\Bigl( S^* \preceq Q_{1-\alpha} \Bigr) \ge \frac{\lceil (n'+1)(1-\alpha) \rceil}{n'+1} \ge 1-\alpha.
$$
Since the prediction set is defined as
$$
\Gamma_{1-\alpha}(X^*) = \{ y \in \mathbb{R}^d : V(X^*,y,\hat{f}) \preceq Q_{1-\alpha} \},
$$
it follows that
$$
\mathbb{P}\Bigl( Y^* \in \Gamma_{1-\alpha}(X^*) \Bigr) = \mathbb{P}\Bigl( S^* \preceq Q_{1-\alpha} \Bigr) \ge 1-\alpha.
$$
\end{proof}

\section{CALIBRATION SCHEMES} \label{app:calib_schemes}
\renewcommand{\arraystretch}{1.3}

\autoref{tab:calib_schemes} summarizes the various calibration schemes in our experiments.

\begin{table}
\centering
\caption{Various conformal calibration schemes.}
\label{tab:calib_schemes}
\small
\begin{tabular}{llcccc}
Method & Description & \begin{tabular}[c]{@{}l@{}}Vector\\ scores?\end{tabular} & \begin{tabular}[c]{@{}l@{}}$\hat F$ \\ estimation\end{tabular} & \begin{tabular}[c]{@{}l@{}}Optimization?\\ \end{tabular} \\ \hline \hline 
\textbf{Independent} & \begin{tabular}[c]{@{}l@{}}univariate calibration applied \\ independently for each target \\ at the $(1-\alpha)^{1/d}$ level\end{tabular} & Y & ECDF & N \\ \hline
\textbf{Scalar score} & 
\begin{tabular}[c]{@{}l@{}}calibration applied to a scalar \\ score defined as the $L_1$-norm\end{tabular}
 & N & ECDF & N \\ \hline
\textbf{\begin{tabular}[c]{@{}l@{}}Empirical\\ copula\end{tabular}} &
\begin{tabular}[c]{@{}l@{}}empirical copula fit on vector \\ scores \cite{messoudi2021copula}\end{tabular}
 & Y & \begin{tabular}[c]{@{}l@{}}ECDF marginals, \\ empirical copula\end{tabular} & Y \\ \hline
\textbf{Plug-in (ours)} & TLL R-vine fit on vector scores & Y & \begin{tabular}[c]{@{}l@{}}ECDF marginals,\\ TLL R-vine\end{tabular} & Y \\ \hline
\textbf{Corrected (ours)} & 
\begin{tabular}[c]{@{}l@{}}One-step correction applied to \\ the above \end{tabular}
& Y & \begin{tabular}[c]{@{}l@{}}ECDF marginals,\\ TLL R-vine\end{tabular} & Y \\ \hline
\end{tabular}
\end{table}


\autoref{tab:synthetic_data_metrics_full} includes a full set of experiments for the penicillin data ($d=3$) with a bigger calibration set ($n=200$), to accommodate methods requiring an extra split of the calibration set. Too few calibration points for the second split $\mathcal{I}_{{\rm cal}, 2}$ will result in infinitely large prediction sets. These methods are labeled ``split.''

For scalar score methods using an extra split, the first split $\mathcal{I}_{{\rm cal}, 1}$ is used to compute the mean and standard deviation of the per-target scores with which to standardize the scores of $\mathcal{I}_{{\rm cal}, 2}$, for usual ranking. This method can be viewed as using a ``fitted target aggregation scheme,'' but still carries the fundamental limitation of scalar scores, of not accounting for the joint correlation structure. Moreover, using the calibration split only to standardize the scores directly may be a waste when the labels can be standardized everywhere, as was done already, at no cost. We observe that the split scalar scores still overcover, as they do not perform joint ranking. 

\begin{table}
\caption{Metrics on the penicillin data ($d=3$, $n=200$). We report mean $\pm$ standard error across ten seeds. Target coverage is 0.9. For efficiency, lower is better. \label{tab:synthetic_data_metrics_full}}
\centering
\small
\begin{tabular}{lcc}
\toprule
\textbf{Method} & {\textbf{Coverage}} & {\textbf{Efficiency} $\downarrow$} \\ \midrule
Independent & $0.95 \pm 0.00$ & $4.0 \pm 0.1$ \\
Scalar score ($L_2$) & $0.92 \pm 0.00$ & $4.3 \pm 0.1$ \\
Scalar score ($L_1$) & $0.92 \pm 0.00$ & $4.6 \pm 0.1$ \\
Scalar score ($L_\infty$) & $0.92 \pm 0.00$ & $4.0 \pm 0.1$ \\
Empirical copula & $0.91 \pm 0.01$ & $3.6 \pm 0.1$ \\
Scalar score ($L_2$), split & $0.93 \pm 0.01$ & $4.4 \pm 0.1$ \\
Scalar score ($L_1$), split & $0.93 \pm 0.01$ & $4.8 \pm 0.2$ \\
Scalar score ($L_\infty$), split & $0.93 \pm 0.01$ & $4.0 \pm 0.1$ \\
Plug-in (ours) & $0.92 \pm 0.00$ & $3.6 \pm 0.2$ \\
Corrected (ours) & $0.90 \pm 0.00$ & $3.6 \pm 0.1$ \\ 
Plug-in, split (ours) & $0.92 \pm 0.03$ & $3.8 \pm 0.2$ \\
Corrected, split (ours) & $0.92 \pm 0.03$ & $3.8 \pm 0.2$ \\
\bottomrule
\end{tabular}
\end{table}

\section{ADDITIONAL EXPERIMENTAL DETAILS} \label{app:additional_experimental_details}

\subsection{Synthetic example} \label{app:synthetic_example}

The penicillin production simulator \citep{liang2021scalable} is a numerical test function that takes in seven input parameters that control penicillin production---culture volume, biomass, concentration, temperature, glucose concentration, substrate feed rate, substrate feed concentration, and proton concentration---and outputs three real-valued target labels: the penicillin yield, production time, and the amount of carbon dioxide byproduct emitted. As shown in \autoref{fig:penicillin_corner}, the target quantities are highly correlated, which lead to correlated residuals in \autoref{fig:penicillin_scores_corner}. The residuals also have upper tails as they most likely contain the 10\% experimental errors.

\begin{minipage}{0.47\textwidth}
    \includegraphics[width=\linewidth]{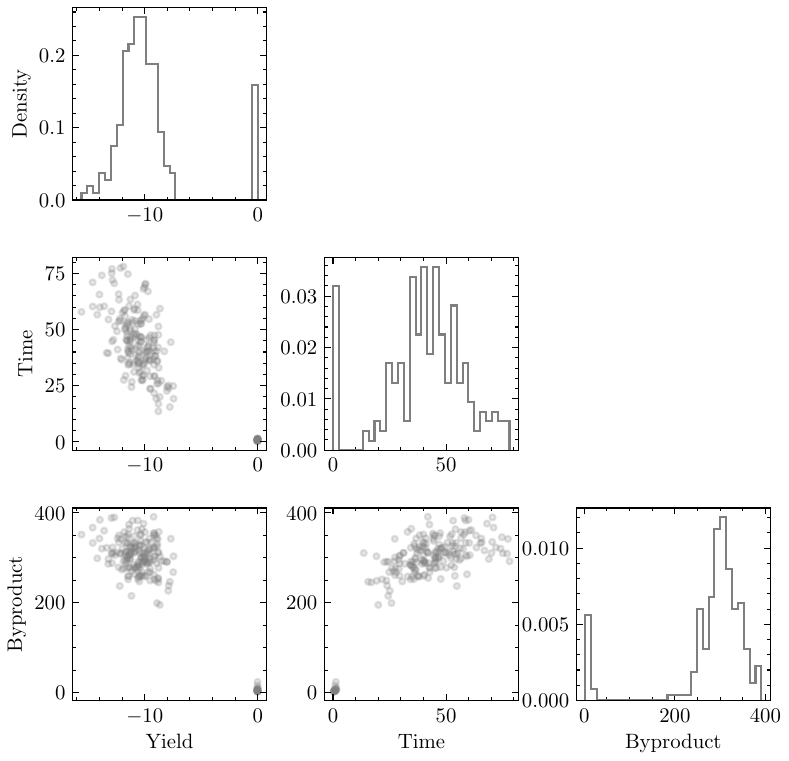}
    \captionof{figure}{\small Highly correlated target quantities.}
    \label{fig:penicillin_corner}
\end{minipage}%
\begin{minipage}{0.47\textwidth}
    \includegraphics[width=\linewidth]{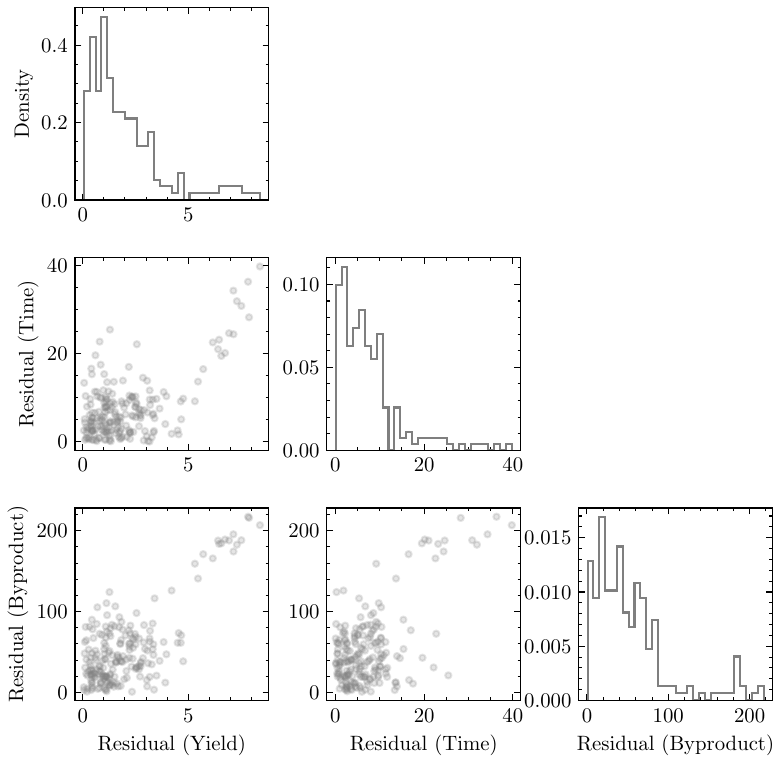}
    \captionof{figure}{\small Highly correlated residuals.}
    \label{fig:penicillin_scores_corner}
\end{minipage}

\begin{figure}
\centering
\includegraphics[width=0.6\linewidth]{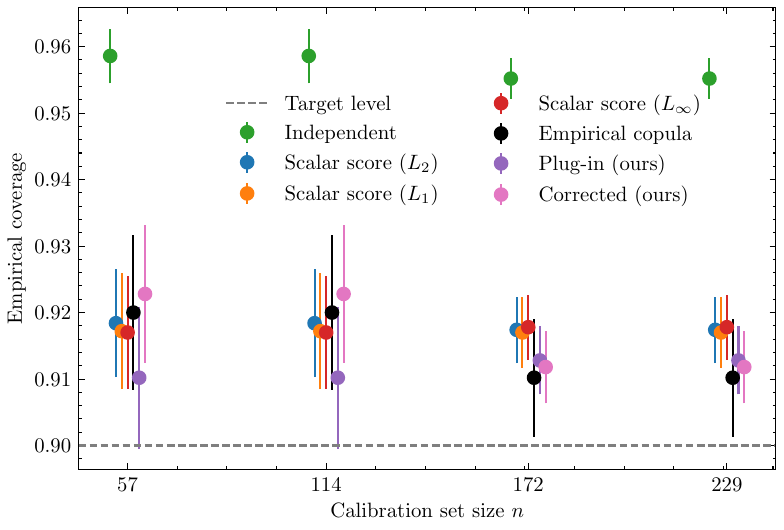}
\caption{\small One-step correction improves the coverage of the plug-in estimate at all calibration-set sizes $n$ for the penicillin ($d=3$) dataset. Error bars are stddev across ten random seeds. \label{fig:cov_vs_n}}
\end{figure}

\autoref{fig:cov_vs_n} plots the mean and standard deviation in coverage for all the methods across ten random seeds as the number of calibration points $n$ increases. Our corrected method consistently yields valid coverage at the 0.91-0.92 level for a target coverage of 0.9. At small $n$, our "plug-in" method has mean coverage of 0.91 with high std (0.1) across ten random seeds, but one-step correction appropriately adjusts the coverage upward. For all $n$, the variances of coverage for our methods are similar to the those of ``independent'' and ``scalar score'' baselines for all 
 and shrink to 0.04 at $n\sim 200$.

\subsection{Real-world data}

\begin{figure}
\centering
\includegraphics[width=0.6\linewidth]{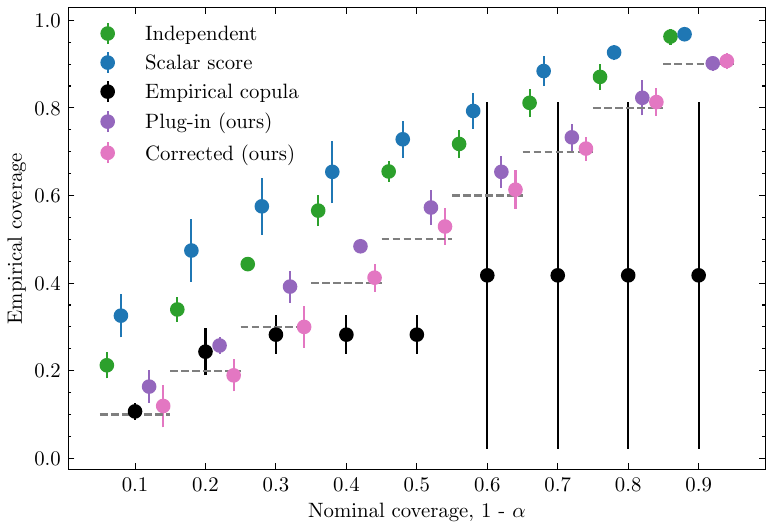}
\caption{\small One-step correction improves the coverage of the plug-in estimate at all target levels for the rf1 ($d=8$) dataset. Error bars are stddev across five random seeds. \label{fig:cov_vs_alpha_rf1}}
\end{figure}

We plot the calibration coverage for varying $\alpha$ for the rf1 dataset ($d=6, n=225$) in \autoref{fig:cov_vs_alpha_rf1}. In line with the observations in \autoref{sec:real_data}, the empirical copula carries high variance due to limited samples in the upper tails close to the target quantile. Its coverage falls substantially below $1-\alpha$ for $1-\alpha \geq 0.4$.

\section{MISSING-AT-RANDOM DATA} 

As $d$ grows, in many practical settings, there may not be many datapoints with all $d$ target variables observed. Using the first two targets ($d=2$) of the penicillin production data, we simulate a missing-at-random (MAR) scenario \citep{rubin1976inference} in which we only observe one target if the observations for another target exceeds a certain threshold. Specifically, we generate the data $\{ (X^{(i)}, Y^{(i)})\}_{i=1}^{420}$ and simulate the missingness variable $\{Z^{(i)}_1\}_{i=1}^{420}$ with $Z_1 = 1$ if $Y_1$ is observed and 0 otherwise. If $Y_1$ is missing, then the score $S_1$ also cannot be evaluated. Variable $Y_2$, on the other hand, is always observed. We define the missingness as follows:
$$
Z^{(i)}_1 = \mathds{1}[Y^{(i)}_2 \leq 50 ],
$$
such that approximately 40\% of $Y_1$ is missing. We split the 420 datapoints into 300 training and 120 calibration instances. The underlying model is a univariate Lasso regressor trained independently on only the observed labels for each target \citep{tibshirani1996regression}. Additionally, we also simulate 500 fully observed test instances.

\begin{figure}
\centering
\includegraphics[width=0.7\linewidth]{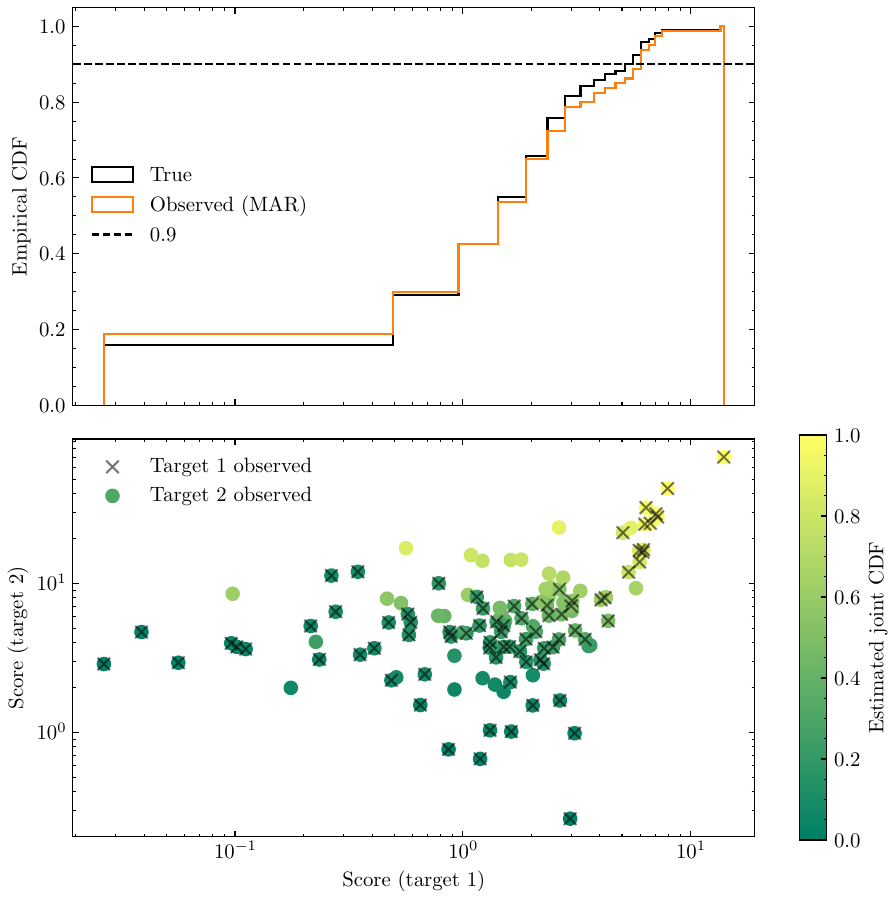}
\caption{\small A scenario in which the target labels are missing at random, such that the labels for target 1 are only observed when target 2 observations exceed a certain value. The empirical CDF of target 1 based on the observations differs from the true CDF of target 1, so only using the fully observed calibration instances results in a biased quantile estimate. \label{fig:mar_example}}  
\end{figure}

Only using the calibration instances for which both target labels are observed results in a biased quantile estimate, as illustrated in \autoref{fig:mar_example}. To include the calibration instances with $Y_1$ missing, we fit a Gaussian copula with missingness imputation \cite{feldman2024nonparametric} using the EM algorithm. As \autoref{tab:mar_example} shows, all the methods without imputation overcover due to the bias introduced by missingness, whereas our methods with the imputation copula achieve the desired coverage of 0.9 and yield the most efficient prediction sets.

\begin{table}
\small
\caption{Metrics on the penicillin data with missing-at-random labels in the training and calibration sets. We report mean $\pm$ standard error across five random seeds. Target coverage is 0.9. For efficiency, lower is better.}
\label{tab:mar_example}
\centering
\begin{tabular}{lcc}
\toprule
\textbf{Method} & {\textbf{Coverage}} & {\textbf{Efficiency} $\downarrow$} \\ \midrule
Independent & $0.96 \pm 0.01$ & $6.7 \pm 0.0$ \\
Scalar score & $0.94 \pm 0.01$ & $7.2 \pm 0.2$ \\
Empirical copula & $0.94 \pm 0.01$ & $6.4 \pm 0.1$ \\
Plug-in (ours) & $0.92 \pm 0.01$ & $6.2 \pm 0.1$ \\
Corrected (ours) & $0.92 \pm 0.01$ & $6.2 \pm 0.1$ \\
\hline
Plug-in with imputation (ours) & $0.90 \pm 0.00$ & $6.0 \pm 0.0$ \\
Corrected with imputation (ours) & $0.91 \pm 0.00$ & $6.1 \pm 0.0$ \\ \bottomrule
\end{tabular}
\end{table}





\section{ALTERNATIVE UNDERLYING ALGORITHMS} \label{app:underlying_predictor}

We include experiments where the underlying model is a conditional density estimator ${\hat f}: \mathcal{X} \rightarrow \mathbb{R}^d \times \mathbb{R}^d_+$ that outputs the mean and variance of the factorized Gaussian predictive distribution. We report results on the penicillin data in \autoref{tab:gaussian_cde_metrics}. The model is a two-layer multi-layer perceptron (MLP) with a hidden dimension of 32, trained to minimize the Gaussian log likelihood,
$$L(\theta) = \sum_{i=1}^{n_{\rm train}} \sum_{j=1}^d \ \log \sigma_j + \frac{\left(Y^{(i)}_j - \mu_j \right)^2}{2 \sigma_j^2},$$
where $\mu_j = \mu_j(X^{(i)}; \theta)$ and $\sigma_j = \sigma_j(X^{(i)}; \theta)$ are the learned mean and standard deviation, respectively, for target $j$ of the Gaussian predictive distribution parameterized by the MLP parameters $\theta$ and $n_{\rm train}$ is the size of the training set. We parameterize $\sigma_j$ as $\log \sigma_j^2$ for numerical stability. 

\begin{table}
\small
\caption{Metrics on the penicillin data ($d=3$) with the Gaussian conditional density estimator as the underlying model. We report mean $\pm$ standard error across five random seeds. Target coverage is 0.9. For efficiency, lower is better.}
\label{tab:gaussian_cde_metrics}
\centering
\begin{tabular}{lcc}
\toprule
\textbf{Method} & {\textbf{Coverage}} & {\textbf{Efficiency} $\downarrow$} \\ \midrule
Independent & $0.96 \pm 0.01$ & $7.9 \pm 0.2$ \\
Scalar score & $0.95 \pm 0.01$ & $6.9 \pm 0.1$ \\
Empirical copula & $0.95 \pm 0.01$ & $7.9 \pm 0.3$ \\
Plug-in (ours) & $0.94 \pm 0.02$ & $7.7 \pm 0.6$ \\
Corrected (ours) & $0.92 \pm 0.01$ & $7.0 \pm 0.2$ \\ \bottomrule
\end{tabular}
\end{table}

\vfill

\end{document}